\documentclass{article}
\usepackage{ijcai21}
\usepackage[a-2b]{pdfx} % pdf/A
\hypersetup{hidelinks}
\usepackage{times}
\usepackage{soul}
\usepackage{url}
\usepackage[utf8]{inputenc}
\usepackage[small]{caption}
\usepackage{graphicx}

\usepackage{algorithm}
\usepackage{algorithmic}
\usepackage[switch]{lineno} 
\usepackage{amsmath,amsfonts,amsthm}
\usepackage{makecell}
\usepackage{multirow}
\usepackage{booktabs}
\usepackage{subcaption}
\usepackage{supertabular} 

\title{Contrastive Losses and Solution Caching for Predict-and-Optimize}

\author{
Maxime Mulamba$^1$
\and
Jayanta Mandi$^1$\and
Michelangelo Diligenti $^{2}$\and
 Michele Lombardi$^3$ \and
 Victor Bucarey$^1$ \and
 Tias Guns$^{1,4}$
\affiliations
$^1$Data Analytics Laboratory, Vrije Universiteit Brussel, Belgium\\
$^2$Department of Information Engineering and Mathematical Sciences, University of Siena, Italy\\
$^3$Dipartimento di Informatica - Scienza e Ingegneria, University of Bologna, Italy\\
$^4$Department of Computer Science, KU Leuven, Belgium\\
\emails
\{maxime.mulamba, jayanta.mandi,victor.bucarey.lopez, tias.guns\}@vub.be,
diligmic@diism.unisi.it,
michele.lombardi2@unibo.it
}

\DeclareMathOperator*{\argmax}{argmax}
\DeclareMathOperator*{\argmin}{argmin}
\DeclareMathOperator*{\NCE}{NCE}
\DeclareMathOperator*{\MAP}{MAP}

\newcommand\michele[1]{\textcolor{SkyBlue}{michele:#1}}

\newcommand\ignore[1]{}

\begin{document}

\maketitle

%\michele{modified abstract (summary + mention of contrastive loss} Thx

\begin{abstract}
Many decision-making processes involve solving a combinatorial optimization problem with uncertain input that can be estimated from historic data.
Recently, problems in this class have been successfully addressed via end-to-end learning approaches, which rely on solving one optimization problem for each training instance at every epoch. In this context, we provide two distinct contributions. First, we use a Noise Contrastive approach to motivate a family of surrogate loss functions, based on viewing non-optimal solutions as negative examples. 
Second, we address a major bottleneck of all predict-and-optimize approaches, i.e. the need to frequently recompute optimal solutions at training time. This is done via a solver-agnostic solution caching scheme, and by replacing optimization calls with a lookup in the solution cache. 
The method is formally based on an inner approximation of the feasible space and, combined with a cache lookup strategy, provides a controllable trade-off between training time and accuracy of the loss approximation. We empirically show that even a very slow growth rate is enough to match the quality of state-of-the-art methods, at a fraction of the computational cost.
\end{abstract}

\section{Introduction}

Many real-life decision-making problems can be formulated as combinatorial optimization problems. 
However, uncertainty in the input parameters is commonplace; an example being the day-ahead scheduling of tasks on machines, where future energy prices are uncertain. 
A \emph{predict-then-optimize}~\cite{elmachtoub2017smart} approach is a widely-utilized industry practice, where first a machine learning (ML) model is trained to make a point estimate of the uncertain parameters and then the optimization problem is solved using the predictions. 

The ML models are trained to minimize prediction errors without taking into consideration their impacts on the downstream optimization problem. This often results in sub-optimal decision performance. A more appropriate choice would be to integrate the prediction and the optimization task and train the ML model using a \emph{decision-focused loss}~\cite{elmachtoub2017smart,wilder2019melding,emir2019investigation}. 
Such predict-\textit{and}-optimize approach is proven to be effective in various tasks~\cite{m2019smart,demirovic2019predict+,ferber2020mipaal}.

Unfortunately, computational complexity and scalability are two major roadblocks for the predict-and-optimize approach involving NP-hard combinatorial optimization problem.
This is due to the fact that an NP-hard optimization problem must be solved and differentiated for each training instance on each training epoch to find a gradient of the optimization task and backpropgating it during model training. 

A number of approaches~\cite{wilder2019melding,ferber2020mipaal,mandi2020interior} consider problems formulated as Integer Linear program (ILP) and
solve and differentiate the relaxed LP using interior point methods. On the other hand, the approaches from \cite{m2019smart} and \cite{vlastelica2019differentiation} are solver-agnostic, because they compute a subgradient using solutions of any combinatorial solvers.

Here we propose an alternative approach, motivated by the literature of noise-contrastive estimation \cite{gutmann2010noise},
which we use to develop a new family of surrogate loss functions based on viewing non-optimal solutions as negative examples. 
This necessitates building a cache of solutions, which we implement by storing previous solutions during training. We provide a formal interpretation of such a solution cache as an inner approximation of the convex-hull of feasible solutions. This is helpful whenever a linear cost vector is optimized over a discrete space. 
Our second contribution is to propose a family of loss functions specific to combinatorial optimization problems with linear objectives.
As an additional contribution, we extend the concept of discrete inner approximation to solver-agnostic approaches. In this way, we are able to overcome the training time bottleneck. 
Finally, we empirically demonstrate that noise-contrastive estimation and solution caching produce predictions at the same quality or better than the state-of-the-art methods in the literature with a drastic decrease in computational times.

\section{Related Work} \label{s:related}
Noise-contrastive estimation (NCE) performs tractable parameter optimization for many models requiring normalization of the probability distribution over a set of discrete assignments. This is a common element of many popular probabilistic logic frameworks like Markov Logic Networks~\cite{richardson2006markov} or Probabilistic Soft Logic~\cite{bach2017hinge}.
More recently, NCE has been at the core of several neuro-symbolic reasoning approaches~\cite{garcez2012neural} like Deep Logic Models~\cite{marra2019integrating} or Relational Neural Machines~\cite{marra2020relational}. We use NCE to derive some tractable formulations of the combinatorial optimization problem. %We here use it to approximate and differentiate over the argmin operator over all feasible solutions of the combinatorial optimization problem.

% MICHELANGELO: I removed the following as we did not really use the logic/learning integration in the paper.
%The integration of logic rules into neural network models can be explored from multiple perspectives. Earlier works~\cite{shavlik1991approach, towell1994} on neuro-symbolic reasoning literature infuse propositional rules by encoding it in euclidean space [see \cite{garcez2012neural} for an extensive review]. 
%More recent works such as semantic-based regularization~\cite{diligenti2017semantic} and Logic Tensor Networks~\cite{donadello2017logic} integrate logic rules by considering continuous relaxation of it.

%DeepProbLog~\cite{manhaeve2018deepproblog} extends ProbLog~\cite{de2007problog}, the  probabilistic programming framework, with the integration of deep neural network to learn probabilistic logical neural models. 
%\cite{xu2018semantic} introduces a similar method to encode probabilistic logic  into  the  loss  function  of  neural  network. 
%Basically, these works in Neuro-Symbolic AI construct a network from a given rule set for decision making. \cite{fischer2019dl2} introduces DL2, a  method  for  deriving  losses from logic rules suitable for gradient base training.
%\cite{wang2019satnet} take a different approach, which learns the logic rules from the data through end-to-end training using differntiable MAXSAT solver.
%Along this line, \cite{donti2017task} apply end-to-end training on a stochastic optimization problem where the loss function captures the number of constraint violations.

Numerical instability is a major issue in end-to-end training as 
implicit differentiation at the optimal point leads to zero Jacobian when optimizing linear functions. \cite{wilder2019melding} introduce end-to-end training of a combinatorial problem by constructing a simpler optimization problem in the continuous relaxation space adding a quadratic regularizer term to the objective. As the continuous relaxation is an outer approximation of the feasible region in mixed integer problems, \cite{ferber2020mipaal} strengthen the formulation by adding cuts. \cite{mandi2020interior} propose to differentiate the homogeneous self-dual formulation, instead of the KKT condition and show its effectiveness.

The Smart Predict and Optimize (SPO) framework introduced by \cite{elmachtoub2017smart} uses a convex surrogate loss based subgradient which could overcome the numerical instability issue for linear problems.
\cite{m2019smart} investigate scaling up the technique for large-scale combinatorial problems using continuous relaxations and warm-starting of the solvers.
Recent work by \cite{vlastelica2019differentiation} is similar to the SPO framework but it uses a different subgradient considering ``implicit interpolation'' of the argmin operator. Both of these approaches are capable of computing the gradient for any blackbox implementation of a combinatorial optimization with linear objective.  \cite{elmachtoub2020decision} extends the SPO framework for decision trees.

In all the discussed approaches, scalability is a major challenge due to the need to repeatedly solve the (possibly relaxed) optimisation problems. In contrast, our contrastive losses, coupled with a solution caching mechanism, do away with repeatedly solving the optimization problem during training and can be applied to other solver agnostic predict-and-optimize methods, too.

\section{Problem Setting}
%\tias{Please proofread the entire paper and never write 'optimization problem' as this may confuse ML people, write 'combinatorial problem' or 'combinatorial optimization problem' always with the combinatorial indicator}
In our setting we consider a combinatorial optimization problem in the form
\begin{equation}
    v^*(c) = \argmin_{v \in V} f(v, c) 
    \label{eq:COP}
\end{equation}
\noindent where $V$ is a set of feasible solutions and $f:V\times \mathcal{C} \rightarrow \mathbb{R}$ is a real valued function. The objective function $f$ is parametric in $c$, the values we will try to estimate. We denote by $v^*(c)$ an optimal solution of \eqref{eq:COP}. Despite the fact that $V$ can be any set, for the rest of the article we will consider the particular case where $V$ is a discrete set, specified implicitly through a set of constraints. This type of sets arise naturally in combinatorial optimization problems, including Mixed Integer Programming (MIP) and Constraint Programming (CP) problems, many of which are known to be NP-complete. % and will allow us to introduce smoothly the solution techniques proposed in this article.

The value of $c$ is unknown but we assume having access to correlated features $x$ and a historic dataset $D = \{(x_i, c_i)\}_{i=1}^n$. One straightforward method to learn $c$ is to find a model $m(\omega, x)$ with model parameters $\omega$ that predicts a value $\hat{c}$. This model can be learned by fitting the data $D$ to minimizing some loss function, as in classical supervised learning approaches. 
% The main drawback of this procedure is to obviate the inner structure of \eqref{eq:COP} leading to poor results in the decision making process. 

In a predict-and-optimize setting, the challenge is to learn model parameters $\omega$, such that, when it is used to provide estimates $\hat{c}$, these predictions lead to an optimal solution of the combinatorial problem with respect to the real values of $c$. In order to measure how good a model is, we compute the regret of the combinatorial optimisation, that is, the difference between the true value of: 1) the optimal solution $v^*(c)$ for the true parameter values; and 2) the optimal solution for the estimated parameter values $v^*(\hat{c})$. Formally, $\mbox{Regret}(\hat{c},c) = f(v^*(\hat{c}), c) -  f(v^*(c), c)$.
% for a given set of parameters $\omega$ the regret is defined by:
% \begin{equation}
% \mbox{Regret}(\hat{c},c) = f(v^*(\hat{c}), c) -  f(v^*(c), c).
% \end{equation}
In case of minimisation problems, regret is always positive and it is $0$ in case optimizing over the estimated values leads either to the true optimal solution or to an equivalent one.

The goal of prediction-and-optimisation is to learn the model parameters $\omega$ to minimize the regret of the resulting predictions, i.e. $\argmin_\omega \mathbb{E}\left[ \mbox{Regret}(m(\omega,x),c) \right]$.
% \begin{equation}
% \argmin_\omega \mathbb{E}\left[ \mbox{Regret}(m(\omega,x),c) \right]. \label{eq:minregret}
% \end{equation}
When using backpropagation as a learning mechanism, regret cannot be directly used as a loss function because it is non-continuous and involves differentiating over the argmin in $v^*(c)$.
Hence, the general challenge of predict-and-optimize is to identify a differentiable and efficient-to-compute loss function $\mathcal{L}^{v^*}$ that takes into account the structure of $f$ and $v^*(\cdot)$ more generally.

Learning over a set of $N$ training instances can be formulated within the empirical risk minimisation framework as 
\begin{align}
\begin{split}
\argmin_\omega \mathbb{E}\left[\mathcal{L}^{v^*}(m(\omega,x),c)\right] \\ \approx \argmin_\omega \frac{1}{N} \sum_{i=1}^N \mathcal{L}^{v^*}(m(\omega, x_i),c_i)
\label{eq:taskloss}
\end{split}
\end{align}
%\begin{algorithm}[t]
%\SetAlgoLined \DontPrintSemicolon
%\SetKwInOut{Input}{Input}\SetKwInOut{Output}{output}
%\SetKwRepeat{Do}{do}{}
%\SetKwRepeat{Repeat}{repeat }{until convergence}
%\SetKwFunction{proc}{}
%\SetKwProg{fwd}{Forward Pass}{}{}
%\SetKwProg{bkwd}{Backward Pass}{}{}
%
%\SetKw{Arg}{Argument:}
%\SetKw{Param}{Parameter:}
%\SetKw{Hyperparam}{Hyperparameters:}
%\SetKw{Solver}{Solver}{}{}
%\SetKwInOut{Input}{input}\SetKwInOut{Output}{output}
%\SetKwRepeat{Do}{do}{}
%\SetKwRepeat{Repeat}{repeat }{until }
%\SetKwFunction{proc}{}
%\Arg{$A, b$}; training data D$ \equiv \{(x_i,c_i)\}_{i=1}^n$
%\Param{$\omega$} \Hyperparam{$\alpha$- learning rate}\\
%Initialize $\omega $\\
%\For{each epochs}{
%\For{each instances}{
%$\hat{c} \leftarrow m (\omega, x)$
%%   $v^*:= \argmin_v f(v,\hat{c})$ ;
%%   \Return $v^*$
%
%Compute $ \mathcal{L}^{v^*}(\hat{c},c)$ by solving $v^*(\hat{c})$
%%  $\frac{\partial \mathcal{L}}{\partial \omega} \leftarrow \frac{\partial \mathcal{L}}{\partial v^*} \frac{\partial v^*}{\partial \hat{c}} \frac{\partial \hat{c}}{\partial \omega}$ ;
%
%$\omega \leftarrow \omega - \alpha \frac{\partial \mathcal{L}^{v^*}}{\partial \omega}$ 
%}
%}
% \caption{Forward and Backward Pass in End-to-end Predict + Optimize}
% \label{Alg}
%\end{algorithm}
\begin{algorithm}[tb]
\caption{Gradient-descent over combinatorial problem}
\label{alg:algorithm}
\textbf{Input}:$A$,$b$; training data D$ \equiv \{(x_i,c_i)\}_{i=1}^n$ \\
\textbf{Hyperparams}: $\alpha$- learning rate, epochs\\[-1em]
%\textbf{Output}: Your algorithm's output
\begin{algorithmic}[1] %[1] enables line numbers
\STATE Initialize $\omega$.
\FOR{each epochs}
\FOR{each instances}
\STATE $\tilde{c} \leftarrow t (\hat{c})$ with $\hat{c} = m (\omega, x)$ \label{alg:ln:pred} %\# transform predictions.
\STATE Obtain $v$ by calling a solver for Eq. \eqref{eq:COP} with $\tilde{c}$ \label{alg:ln:solve}  % Compute $ \mathcal{L}^{v^*}(\hat{c},c)$ by solving $v^*(\hat{c})$
\STATE $\omega \leftarrow \omega - \alpha \frac{\partial \mathcal{L}^{v}}{\partial \tilde{c}} \frac{ \partial \tilde{c}}{\partial \omega}$ \# backpropagate (sub)gradient
\ENDFOR
\ENDFOR
\end{algorithmic}
\label{Alg}
\end{algorithm}
\subsection{Gradient-Descent Decision-Focused Learning}
Algorithm~\ref{Alg} depicts a standard gradient descent learning procedure for predict-and-optimize approaches.
For each epoch and instance, it computes the predictions, optionally transforms them on Line~\ref{alg:ln:pred}, calls a solver to compute $v^*(\tilde{c})$, and updates the trainable weights $\omega$ via standard backpropagation for an appropriately defined gradient $\partial \mathcal{L}^{v} / \partial c$.

To overcome both the non-continuous nature of the optimisation problem $v^*(c)$ and the computation time required, a number of works replace the original task $v^*$ by a continuous relaxation $g^*$ and solve and \textit{implicitly} differentiate over $\mathcal{L}^{g^*}$, considering a quadratic \cite{wilder2019melding} or  log-barrier \cite{mandi2020interior} task-loss. In these cases, $t(\hat{c}) = \hat{c}$.

Other approaches are solver-agnostic and do the implicit differentiation by defining a subgradient for $\partial \mathcal{L}^{v} / \partial c$. In case of SPO+ loss \cite{elmachtoub2017smart}, the subgradient is $v^*(c) - v^* ( 2\hat{c} -c )$, involving $t(\hat{c}) = ( 2\hat{c} -c )$. In case of Blackbox differentiation \cite{vlastelica2019differentiation}, the solver is called twice on Line~\ref{alg:ln:transform} of Alg~\ref{Alg} and the subgradient is an interpolation of $\mathcal{L}^{v}$ around $\hat{c}$, where the interpolation is between $v^*(\hat{c})$ and its perturbation $v^*( \hat{c} + \lambda c )$. 
% by perturbing controlled by a parameter $\lambda$.
% The resulting subgradient takes the form $\frac{v^*( \hat{c} + \lambda c ) - v^*(\hat{c})}{\lambda}$. 
% The transformation is a perturbation that is needed \tias{because...? Jay?}

%Secondly, due to the argmin of $v^*$ there are no straightforward loss functions to use, as $\partial \mathcal{L}^{v^*} / \partial c $ involves \emph{argmin} differentiation.
In all those cases, in order to find the (sub)gradient,
% $\partial \mathcal{L}^{v^*} / \partial c $, 
the optimization problem $v^*(c)$ must be solved repeatedly for each instance.
In the next section, we present an alternative class of \textit{contrastive} loss functions that has, to the best of our knowledge, \emph{not} been used before for predict-and-optimize problems. These loss functions can be differentiated in closed-form and do not require solving a combinatorial problem $v^*(c)$ for every instance.

\section{A Contrastive Loss for Predict-and-Optimize} \label{sect-contrastive_loss}
Probabilistic models define a parametric probability distributions over the feasible assignments, and Maximum Likelihood Estimation can be used to find the distribution parameters making the observed data most probable under the model~\cite{kindermann1980markov}. In particular, the family of exponential distributions emerges ubiquitously in machine learning, as it is the required form of the optimal solution of any maximum entropy problem~\cite{berger1996maximum}.

We now propose an exponential distribution that fits the optimisation problem of Eq.~\eqref{eq:COP}. Let $v \in V$ be the space of feasible output assignments $V$ for one example $x$. Then, we define the following exponential distribution over $V$:
\begin{align}\label{eq:exponential}
%    p(v_1, &\ldots, v_n |m(\omega, x_1), \ldots, m(\omega, x_n)) =\nonumber\\
%    &= \displaystyle\prod_i p(v^*_i(c)|m(\omega, x_i)) = \label{eq:exponential}\\
%    &= \frac{1}{Z} \exp\Big(\displaystyle\sum_i -f(v_i, m(\omega, x_i)) \Big)
p(v|m(\omega, x)) = \frac{1}{Z} \exp\Big(-f(v, m(\omega, x)) \Big)
\end{align}
%where $n$ is the number of examples and
the partition function Z normalizes the distribution over the assignment space $V$:
\[
%Z= \displaystyle\sum_i \sum_{v^\prime_i \in V}  \exp\Big(-f(v^\prime_i, m(\omega, x_i)) \Big) \ .
Z= \displaystyle \sum_{v^\prime\in V}  \exp\Big(-f(v^\prime, m(\omega, x)) \Big) \ .
\]
By construction, if $v^*(m(\omega, x))$ is the minimizer of Eq.~\ref{eq:COP} for an instance $x$, it will maximize Eq.~\eqref{eq:exponential} and vice versa. We can use this to fix the solution to $v=v^*(c)$ with $c$ being the true costs, and learn the network weights $\omega$ that maximize the likelihood $p(v^*(c)|m(\omega, x))$. This corresponds to learning an $\omega$ that makes the intended true solution $v^*(c)$ be the best scoring solution of Eq.~\ref{eq:exponential} and hence of $v^*(m(\omega, x))$, which is the goal of prediction-and-optimisation.
% (Eq. \eqref{eq:minregret}).
In the following, these definitions will be implicitly extended over all training instance ($x_i, c_i$).

A main challenge of working with this distribution is that computing the partition function $Z$ requires iterating over all possible solutions $V$, which is intractable for most combinatorial optimization problems.

\subsection{Noise-Contrastive Estimation} Learning over this distribution without a direct evaluation of $Z$ can be achieved by using \emph{Noise Contrastive Estimation} (NCE)~\cite{mikolov2013distributed}. The key idea there is to work with a small set of \textit{negative} samples. To apply NCE in this work, we will use as negative samples the solutions that are different from the target solution $v^\star$, that is any subset $S \subset (V\setminus v^\star)$ of feasible solutions. 

Such an NCE approach avoids a direct evaluation of $Z$ and instead maximizes the separation of the probability of the optimal solution ${v}^\star_i = v^*(c_i)$ for $x_i$ from the probability of a sample of the non-optimal ones (the `noise' part). It is expressed as a maximization of the product of ratios between the optimal solution $v^\star_i$ and the negative samples $S$:
\begin{align}
 \argmax_\omega \log \prod_i \prod_{v^s \in S}  &\frac{p\Big(v^\star_i|m(\omega, x_i) \Big)}{p\Big(v^s|m(\omega, x_i) \Big)} =\label{eq:prob_pol} \\
 = \argmax_\omega \sum_i \sum_{v^s \in S} &\Big(-f(v^\star_i,m(\omega, x_i) ) -log(Z) \nonumber\\
 &+f(v^s,m(\omega, x_i)) +log(Z)\Big) \nonumber \\
 = \argmax_\omega \sum_i \sum_{v^s \in S} &\Big(f(v^s,m(\omega, x_i))-f(v^\star_i,m(\omega, x_i) )  \Big).\nonumber
 \end{align}
\noindent By changing the sign to perform loss minimization, this leads to the following NCE-based loss function:
\begin{align} 
		\mathcal{L}_{\NCE} = \sum_{i} \sum_{v^s \in S} \Big(f\big(v^\star_i, m(\omega, x_i) \big) -  f\big(v^s, m(\omega, x_i \big) \Big)
		\label{eq:l1}
\end{align}
which can be plugged directly into Algorithm~\ref{alg:algorithm}. During differentiation, both $v^\star_i$ and  $v^s$ will be treated as constants- the first since it effectively never changes, the second since it will be computed in the forward pass on line~\ref{alg:ln:solve} in Alg.~\ref{alg:algorithm}. As a side effect, automatic differentiation of Eq.~\ref{eq:l1} will yield a subgradient rather than a true gradient, as is common in integrated predict-and-optimize settings. In section~\ref{sect:5}, we will discuss how to create the sample $S$.
%and the training examples are assumed to be independent given the model predictions.

\subsection{MAP Estimation} Self-contrastive estimation~\cite{goodfellow2014distinguishability} is a special case of NCE where the samples are drawn from the model. A simple but very efficient self-contrastive algorithm takes a single sample, which is the {\it Maximum A Posteriori} (MAP) assignment, i.e. the most probable solution for each example according to the current model $m(\omega, \cdot)$. %ThiFs corresponds to the following approximation of the log-likelihood for a given example:
%$$
%\begin{array}{l}
%\log p(v|m(\omega, x)) = -f(v,m(\omega, x)) - \log Z \\
%~~~~~~~~\approx-f(v,m(\omega, x)) - \displaystyle\max_{v' \in V}[ -f(v',m(\omega, x))]\\
%~~~~~~~~=-f(v,m(\omega, x)) + \displaystyle\min_{v' \in V}[f(v',m(\omega, x))]\\
%~~~~~~~~=-f(v,m(\omega, x)) + f(v^*(m(\omega, x)),m(\omega, x))
%\end{array}
%$$
%where $v^*(m(\omega, x) = \min_{v' \in V}[f(v',m(\omega, x))]$ is the optimal solution computed for the predicted cost vector $m(\omega, x)$.
Therefore , the MAP assignment approximation trains the weights $\omega$ as:
\[
\displaystyle\argmax_\omega \displaystyle\sum_i \left[ - f(v^\star_i, m(\omega, x_i)) + f(\hat{v}^\star_i, m(\omega, x_i)) \big] \right]
\]
with $\hat{v}^\star_i = \argmin_{v' \in S}[f(v',m(\omega, x_i))]$ being the MAP solution for the current model.
With a sign change to switch optimization direction, this translates into the following loss variant: %where the predictor parameters are$optimized to minimize the maximum value of $f\big(v^\star_i, m(\omega, x_i)\big) - f\big(v^s, m(\omega, x_i) \big)$ within the solution sample $S$: % with $v^s$ the result of calling a solver to solve Eq. \eqref{eq:COP}: \michelangelo{I do not understand this last sentence "with $v^s$ the result of calling a solver to solve Eq. \eqref{eq:COP}" this depends on the growing strategy that it is not explained yet, so it should be dropped}
\begin{align} 
		\mathcal{L}_{\MAP} = \displaystyle\sum_{i}
		\big[ f(v^\star_i, m(\omega, x_i)) - f(\hat{v}^\star_i, m(\omega, x_i) ) \big]
		\label{eq:nce_base}
%		\\
%		\mathcal{L}^v_{NCE\_MAP}(c_i,\hat{c}) = \sum_{i} \Big( f\big(v^\star(c_i), \hat{c}) - f\big(v^s, \hat{c} \big) \Big)
%		\label{eq:l3_t}
\end{align}

%where $\hat{v}^*_i = \displaystyle\argmin_{v^s\in S} \big[f(v^s, m(\omega, x_i))  \big]$.
%with $v^s = v^\star(\hat{c})$.

\subsection{Better Handling of Linear Cost Functions} 
%\michele{ATTACH TO THE PART ABOVE}
%\michelangelo{I changed the above to linked with yours, let me know if it makes sense to you (commenting your first equation below as it is the same as (5) above now.}

%The final form of the NCE loss is:
%\begin{align} 
%  \mathcal{L}_{MAP} = \sum_{i} f\left(v^\star_i, m(\omega, x_i)\right) - f\left(\hat{v}_i^*,  m(\omega, x_i) \right) 
%		\label{eq:nce_base}
%\end{align}
%with:
%$$
%\hat{v}^*_i = v^*\left(m(\omega, x_i)\right)
%$$
\ignore{
Each term in the summation from Equation~\eqref{eq:nce_base} is non negative, since $\hat{v}^\star_i$ is by definition the optimal solution with the predicted cost vector $m(\omega, x_i)$.
}

The losses can be minimized by either matching the true optimal solution (the intended behavior), or by making $f\left(v^\star_i, m(\omega, x_i)\right)$ and $f\left(\hat{v}^\star_i,  m(\omega, x_i)\right)$ identical by other means.
% producing bogus costs for which $f\left(v^\star_i, m(\omega, x_i)\right)$ and $f\left(\hat{v}^\star_i,  m(\omega, x_i)\right)$ become identical.
%The latter behavior is undesirable, since it leads in our context to useless training outcomes. 
%
For example, with a linear cost function $f(v,c) = c^T v$, Eq.~\ref{eq:l1} translates to:
\begin{align} 
  \mathcal{L}_{\NCE} = \sum_{i} \sum_{v^s \in S} m(\omega, x_i)^T (v^\star_i
  - v^s) \label{eq:nce_base_linear}
\end{align}
which can be minimized by predicting null costs, i.e. $m(\omega, x_i) = 0$. %\tias{This is true also for non-linear?? if preds=0, function is constant and all solutions are equivalent? well... if this is the entire objective function. Maybe that is why other methods perturb the predictions and not the loss? then also compatible with non-linear loss? [random thought]}
%We propose to fix this issue by adding a constant offset to $\mathcal{L}_{NCE}$, equal to $-c_i^T (v^\star_i - \hat{v}^\star_i)$, so that all original optima are preserved\tias{Are they? I am missing something in this explanation...}:
%In this case, the issue can be fixed by adjusting $\mathcal{L}_{MAP}$ as follows:
To address this issue, we introduce a variant of Eq.~\ref{eq:l1}, where we replace the $\hat{c_i}$ term in the loss with $\big( \hat{c_i} -c_i \big)$. The modification amounts to adding a constant (so that all optimal solutions are preserved), and can be viewed as a regularization term that keeps $\hat{c}$ close to $c$. Thus we get:
\begin{align}
			\mathcal{L}_{\NCE}^{(\hat{c}-c)} &=  \sum_{i} \sum_{v^s \in S} \Big( (m(\omega, x_i) - c_i)^T (v^\star_i - v^s) \Big) 
\end{align}
where $\hat{c}$ (in the loss name) is a shorthand for $m(\omega, x_i)$. Note that we do not perturb the predictions prior to computing $\hat{v}^\star_i$, but only in the loss function.

The loss is still guaranteed non-negative, since $v^\star_i$ is by definition the best possible solution with the cost vector $c_i$. Eq.~\ref{eq:nce_base_linear} can no longer be minimized with a null cost vector;
instead, the loss can only be minimized by having the predicted costs $\hat{c}_i$ match the true costs $c_i$ or, and implied by that, having the solution with the estimated parameters $\hat{v}^\star_i$ match the true optimal solution $v^\star_i$.
The same approach applied to the MAP version leads to:
\begin{align} 
  \mathcal{L}_{\MAP}^{(\hat{c}-c)} = \sum_{i} (m(\omega, x_i) - c_i)^T (v^\star_i - \hat{v}^\star_i) \label{eq:nce_base_linear2}
\end{align}

\section{Negative Samples and Inner-approximations}\label{sect:5}
%\maxime{ Transition from MAP to ALL variant by stating how we can have a stronger estimation of noise by sampling more solutions.   }

%Discrete solution caching approximations

%\maxime{If we need optimal solution of each instance, we can precompute them from ground truth and them. Now we have a caching containing optimal solutions that we can use instead of calling the solver repetitively. Hence the solution caching section. }

%\maxime{Replace 'solution caching' with \texttt{Inner approximation}}

%From the complexity of estimating the partition function $Z$ it is common in NCE to use a set of \textit{noise samples} $S$ such that $Z\approx \displaystyle\max_{v'\in V} -f(v', m(\omega, x)) \approx \max_{v'\in S} -f(v', m(\omega, x))$.  Since the feasible set $V$ does not change, the partition function remains the same over all the instances. Hence, we can {\it cache} all solutions obtained when \eqref{eq:COP} is solved for a fixed cost vector $c$. For an $|S| << |V|$ of reasonable size, computing $v^*(c)$ now corresponds to computing efficiently $\argmin_{v \in S} f(v, c)$ for a tractable solution cache $S$, which can trivially be done by iterating over the solutions one by one.

\subsection{Negative Sample Selection}
The main question now is how to select the `noise', i.e. the negative samples $S$. The only requirement is that any example in $S$ is a feasible solution, i.e. $S \subseteq V$.
Instead of computing multiple feasible solutions in each iteration, which would be \textit{more} costly, we instead propose the pragmatic approach of storing each solution found when calling the solver on Line~\ref{alg:ln:solve} of Alg.~\ref{alg:algorithm} in a solution cache. As training proceeds, the solution cache will grow each time a new solution is found, and we can use this solution cache as negative samples $S$.

\begin{figure}[b]
    \centering
    \includegraphics[scale=1]{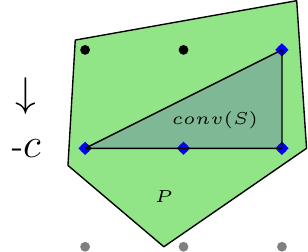}
    \caption{Representation of a solution cache (blue) and the continuous relaxation (green) of $V$.}
    \label{fig:solution_pool}
\end{figure}

While pragmatic, we can also interpret this solution cache $S$ from a combinatorial optimisation perspective: if $S$ would contain all possibly optimal solutions (for linear cost functions), it would represent the convex hull of the entire feasible space $V$. When containing only a subset of points, that is, a subset of the convex hull, it can be seen as an \textit{inner} approximation of $V$. This in contrast to continuous relaxation that relax the integrality constraints, which are commonly used in prediction-and-optimisation today, that lead to an \textit{outer} approximation. 
The inner approximation has the advantage of having more information about the structure of $V$. %For integer linear programs, if the solution cache includes the points on the convex-hull of $V$, it will perfectly represent the set of optimal solutions in $V$. 
This is depicted in Figure \ref{fig:solution_pool} where a solution cache $S$ is represented by blue points, and the set $V$ of feasible points is the union of black and blue points. The continuous relaxation of this set depends on the formulation, that is, the precise set of inequalities used to represent $V$, for example, the green part, which clearly is an outer approximation of the convex-hull of $V$. The convex-hull of the solution cache is represented as $conv(S)$ and it is completely included in $conv(V)$ in contrast to the outer approximation. % an inner approximation.
%By optimizing a linear function $c^t v$ over $V$, as in the picture, we can see that the outer approximation returns the bottom vertex of set $P$ while the inner approximation returns the optimal solution in $S$ that coincides with the set of optimal solutions in $V$. Better continuous relaxations are the ones that are tighter wrt the convex hull of $V$. Even for combinatorial optimization problems that are solvable in polynomial-time, these formulations can include an exponential amount of constraints, as in the case of matching in general graphs.
\subsection{Gradient-descent with Inner Approximation}
The idea that caching the computed solutions results in an inner approximation, is not limited to noise-contrastive estimation. As $S$ becomes larger we can expect the inner approximation to become tighter, and hence we can solve the inner approximation instead of the computationally expensive full problem. Because the inner approximation is a reasonably sized list of solutions, solving it simply corresponds to a linear-time argmin over this list!

\begin{algorithm}[tb]
\caption{Gradient-descent with inner approximation}
\label{alg:algorithm_grow}
\textbf{Input}:$A$,$b$; training data D$ \equiv \{(x_i,c_i)\}_{i=1}^n$ \\
\textbf{Hyperparams}: $\alpha$ learning rate, epochs, $p_{solve}$ \\[-1em]
\begin{algorithmic}[1] %[1] enables line numbers
\STATE Initialize $\omega$
\STATE Initialize $S= \{v^*(c_i) | (x_i, c_i) \in D \}$ \label{alg:ln:inits}
\FOR{each epochs}
\FOR{each instances}
\STATE $\tilde{c} \leftarrow t (\hat{c})$ with $\hat{c} = m (\omega, x)$ \label{alg:ln:transform}
\IF{random$() < p_{solve}$} \label{alg:ln:growth}
    \STATE Obtain $v$ by calling a solver for Eq. \eqref{eq:COP} with $\tilde{c}$
    \STATE $S \leftarrow S \cup \{v\}$
\ELSE
    \STATE $v = \argmin_{v' \in S}(f(v',\tilde{c}))$ \//\// simple argmin \label{alg:ln:lookup}
\ENDIF
\STATE $\omega \leftarrow \omega - \alpha \frac{\partial \mathcal{L}^{v}}{\partial \tilde{c}} \frac{ \partial \tilde{c}}{\partial \omega}$ \# backpropagate (sub)gradient
\ENDFOR
\ENDFOR
\end{algorithmic}
\label{Alg2}
\end{algorithm}

Alg.~\ref{alg:algorithm_grow} shows the generic algorithm. In comparison to Alg.~\ref{alg:algorithm} the main difference is that on Line~\ref{alg:ln:inits} we initialise the solution pool, for example with all true optimal solutions; these must be computed for most loss functions anyway. On Line \ref{alg:ln:growth} we now first sample a random number between 0 and 1, and if it is below $p_{solve}$, which represents the probability of calling the solver, then the expensive solver is called and the solution is added to the cache if not yet present, otherwise an argmin of the cache is done.

\begin{table}[t]
	\centering
	%\small
% 	\resizebox{\textwidth}{!}{
		\begin{tabular}{lclccccclllcccc}
			\toprule
			&                 \makecell{Knapsack\\60} & \makecell{Knapsack\\120} & \makecell{Knapsack\\180}  \\ \toprule
			$\mathcal{L}_{\NCE}$  $\hat{c}$        &   912(21)   & 760(12)      &   2475(45)   \\[0.2em]
			$\mathcal{L}_{\NCE}$  $(\hat{c}-c) $   &  1024(66)   & 770(15)      &  2474 (40)  \\[0.2em] \midrule
			$\mathcal{L}_{\MAP}$  $\hat{c}$ &      1277(555)  & 912(9)       &    491(8)  \\[0.2em]
			$\mathcal{L}_{\MAP}$  $(\hat{c}-c) $   &   \textbf{764 (2)}   & \textbf{562(1) }      &    \textbf{327(1)  }   \\[0.2em] \midrule
			Two-stage &   989 (14) & 1090 (27) & 433 (12)\\[0.2em]

\bottomrule 
		\end{tabular}
% 	}
	\caption{Comparison among Contrastive loss variants on Knapsacks (Average and standard deviation of regret on test data) }
	\label{table:Q1_knap}
\end{table}

\begin{table}[bt]
	\centering
	%\small
% 	\resizebox{\textwidth}{!}{
		\begin{tabular}{lclccc}
			\toprule
			&                 \makecell{Energy\\1}    &   \makecell{Energy\\2}   &  \makecell{Energy\\3}    \\ \toprule
			$\mathcal{L}_{\NCE}$  $\hat{c}$        &     \makecell{45847\\(780)}   &  \makecell{\textbf{27633}\\\textbf{(214)}}  & \makecell{18789\\(194)}  & \\[0.2em]
			$\mathcal{L}_{\NCE}$  $(\hat{c}-c) $   &  \makecell{45834\\(1657)}   &  \makecell{28994\\(659)}  & \makecell{18768\\(406)}  & \\[0.2em] \midrule
			$\mathcal{L}_{\MAP}$  $\hat{c}$ &     \makecell{104496\\(18109)} & \makecell{50897\\(20958)} & \makecell{32180\\(8382)} & \\[0.2em]
			$\mathcal{L}_{\MAP}$  $(\hat{c}-c) $   &    \makecell{\textbf{41236}\\\textbf{(66)}}  & \makecell{27734\\(267)}  &  \makecell{\textbf{17507}\\\textbf{(42) }} & \\[0.2em] \midrule
			Two-stage &  \makecell{43384\\(376)} & \makecell{31798\\(781)} & \makecell{23423\\ (893)} & \\[0.2em]

\bottomrule 
		\end{tabular}
% 	}
	\caption{Comparison among Contrastive loss variants  on Energy scheduling (Average and standard deviation of regret on test data) }
	\label{table:Q1_scheduling}
\end{table}

\begin{table}[t]
	\centering
	%\small
% 	\resizebox{\textwidth}{!}{
		\begin{tabular}{lclccccclllcccc}
			\toprule
			&                 \makecell{Matching\\10} & \makecell{Matching\\25} & \makecell{Matching\\50} & \\ \toprule
			$\mathcal{L}_{\NCE}$  $\hat{c}$      & 3702	(64) & 3696	(76) & 3382	(49) & \\[0.2em]
			$\mathcal{L}_{\NCE}$  $(\hat{c}-c) $    & \textbf{3618	(81)} & \textbf{3674	(48)} & \textbf{3376	(73)} & \\[0.2em] \midrule
			$\mathcal{L}_{\MAP}$  $\hat{c}$  & 3708	(88) & 3700	(23) & 3444	(74) & \\[0.2em]
			$\mathcal{L}_{\MAP}$  $(\hat{c}-c) $    & 3732	(85) & 3712	(86) & 3402	(66) & \\[0.2em] \midrule
			Two-stage &  3700 (42)  & 3712 (59) & 3440 (36)\\[0.2em]
\bottomrule 
		\end{tabular}
% 	}
	\caption{Comparison among Contrastive loss variants on Diverse Bipartite Matching (Average and standard deviation of regret on test data) }
	\label{table:Q1_matching}
\end{table}

\ignore{
The solution caching $S$ can be initialised in many different ways. Based on the idea of having points in the solution in the region of where the optimal points lay, we propose to initialise $S$ by the true optimal solutions on the training data $D$:
\begin{equation}
    S= \{v^*(c_i) | (x_i, c_i) \in D \} \label{eq:s_init}
\end{equation}
This is done before training starts, e.g. on line 2 of Algorithm~\ref{Alg} and \ref{Alg2}.
}

Note how the probability of solving $p_{solve}$ has an efficiency-accuracy trade-off: more solving is computationally more expensive but leads to better approximations of V. The approach of Alg.~\ref{alg:algorithm} corresponds to $p_{solve}=1$.

This inner-approximation caching approach can be used for any decision-focused learning method that calls an external solver, such as SPO+ method in \cite{m2019smart} and its variants~\cite{elmachtoub2020decision}, Blackbox solver differentiation of ~\cite{vlastelica2019differentiation} and our two contrastive losses $\mathcal{L}_{\MAP}$ and $\mathcal{L}_{\NCE}$.
\section{Empirical Evaluation}
In this section we answer the following research questions:
\begin{figure*}[ht]
\centering
\begin{subfigure}[b]{0.28\textwidth}
    \includegraphics[width=\linewidth]{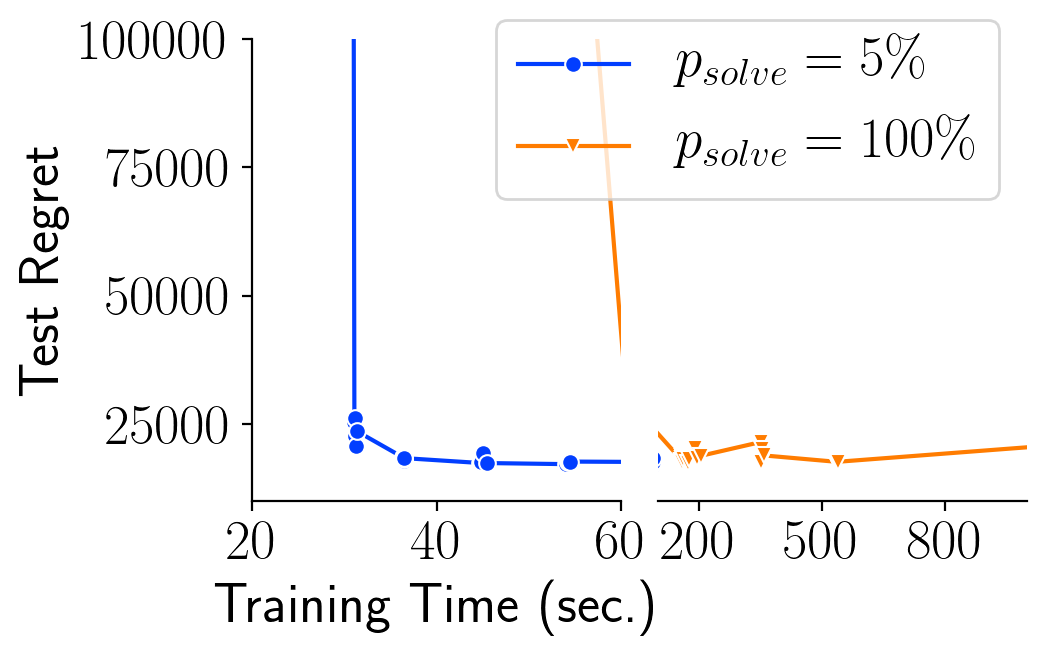}
    \caption{NCE}
    \label{fig:NCE}
  \end{subfigure}
  ~
  \begin{subfigure}[b]{0.28\textwidth}
    \includegraphics[width=\linewidth]{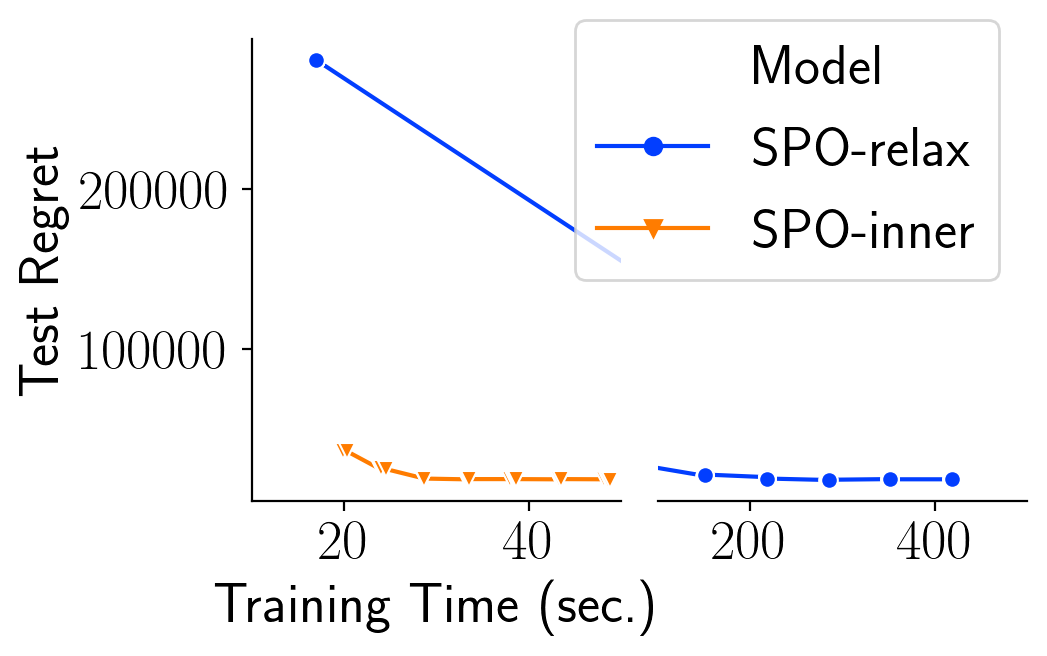}
    \caption{SPO+}
    \label{fig:SPO}
  \end{subfigure}
  ~
  \begin{subfigure}[b]{0.28\textwidth}
    \includegraphics[width=\linewidth]{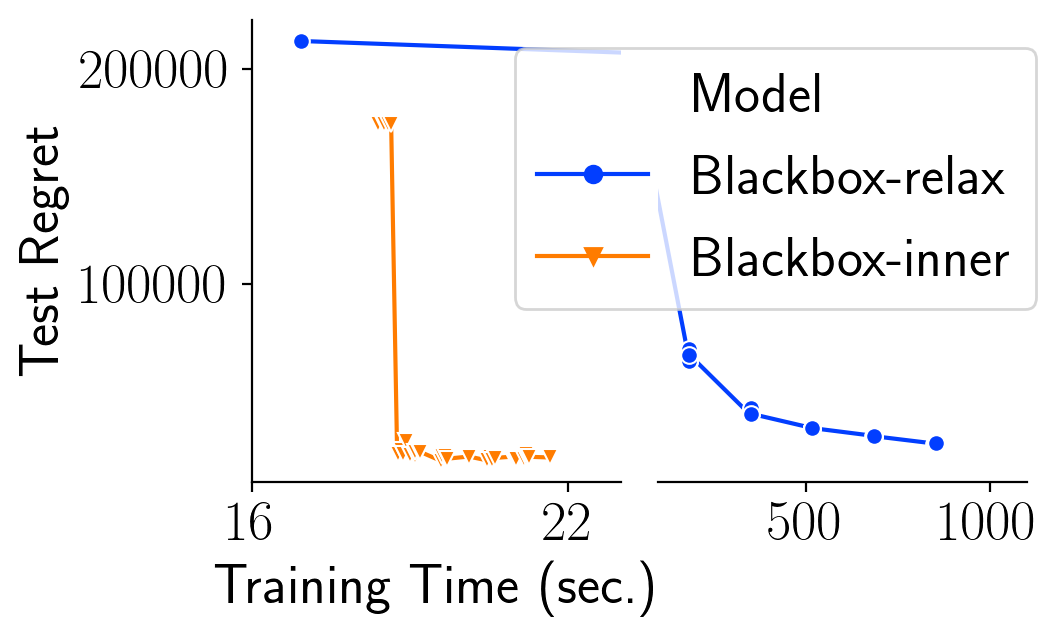}
    \caption{Blackbox}
    \label{fig:BB}
  \end{subfigure}
  \caption{Comparison of learning curves with/without the inner approximation with $p_{solve}=5\%$ for Energy-3.}% we show inner approximation with $p_{solve}=5\%$ reduces the training time without impacting the regret in case of NCE, SPO+ and Blackbox}
\end{figure*}
\begin{itemize}
\itemsep0em 
    \item[\bf Q1] What is the performance of each task loss function in terms of expected regret?
    \item[\bf Q2] How does the growth of the solution caching impact on the solution quality and efficiency of the learning task?
    \item[\bf Q3] How do other solver-agnostic methods benefit from the solution caching scheme? 
    \item[\bf Q4] How does the methodology outlined above perform in comparison with the state-of-the-art algorithms for decision-focused learning?
\end{itemize}
\begin{figure*}[ht]
\centering
\begin{subfigure}[b]{0.28\textwidth}
    \includegraphics[width=\linewidth]{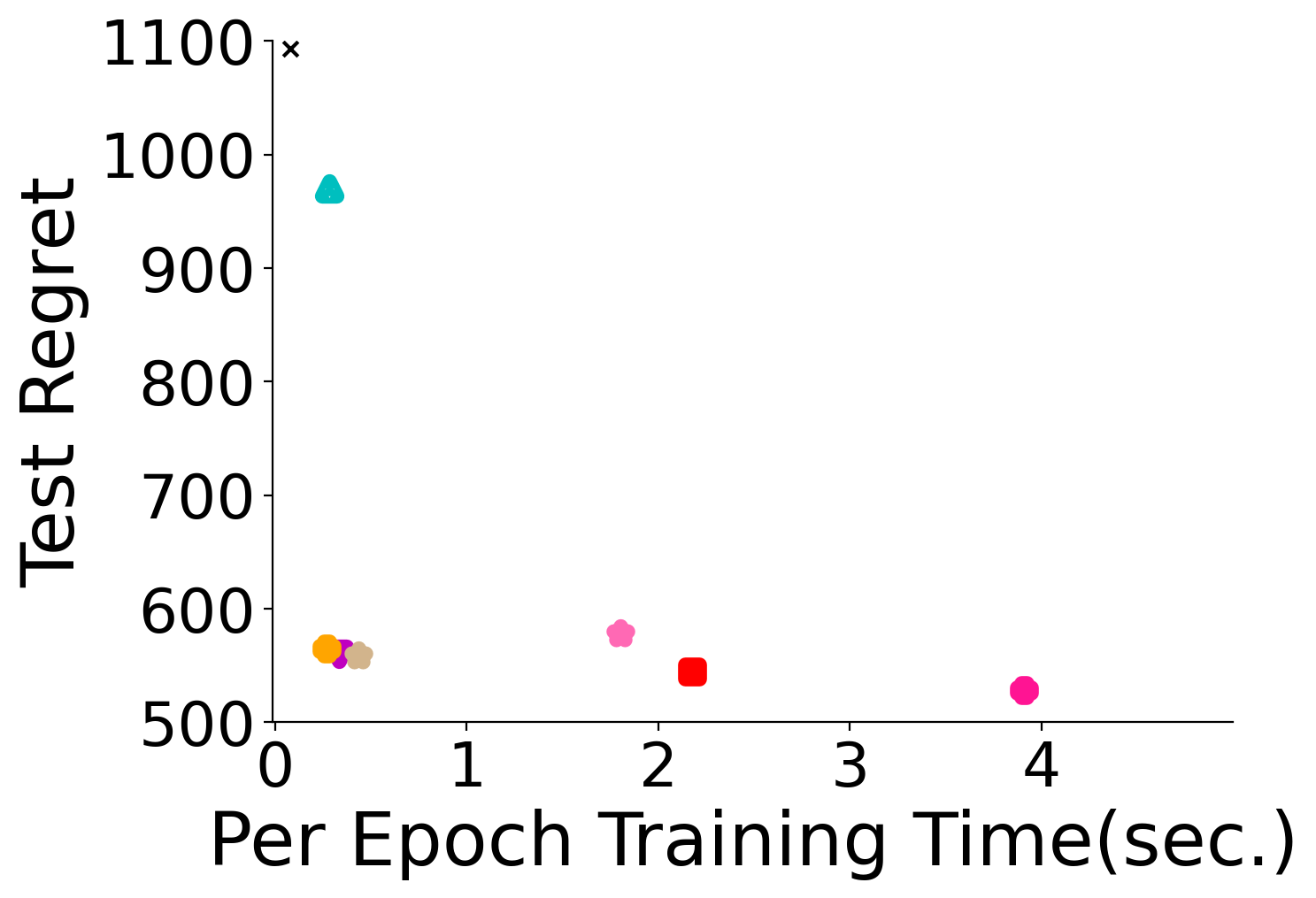}
    \caption{Knapsack-120}
    \label{fig:knapsack}
  \end{subfigure}
  ~
  \begin{subfigure}[b]{0.28\textwidth}
    \includegraphics[width=\linewidth]{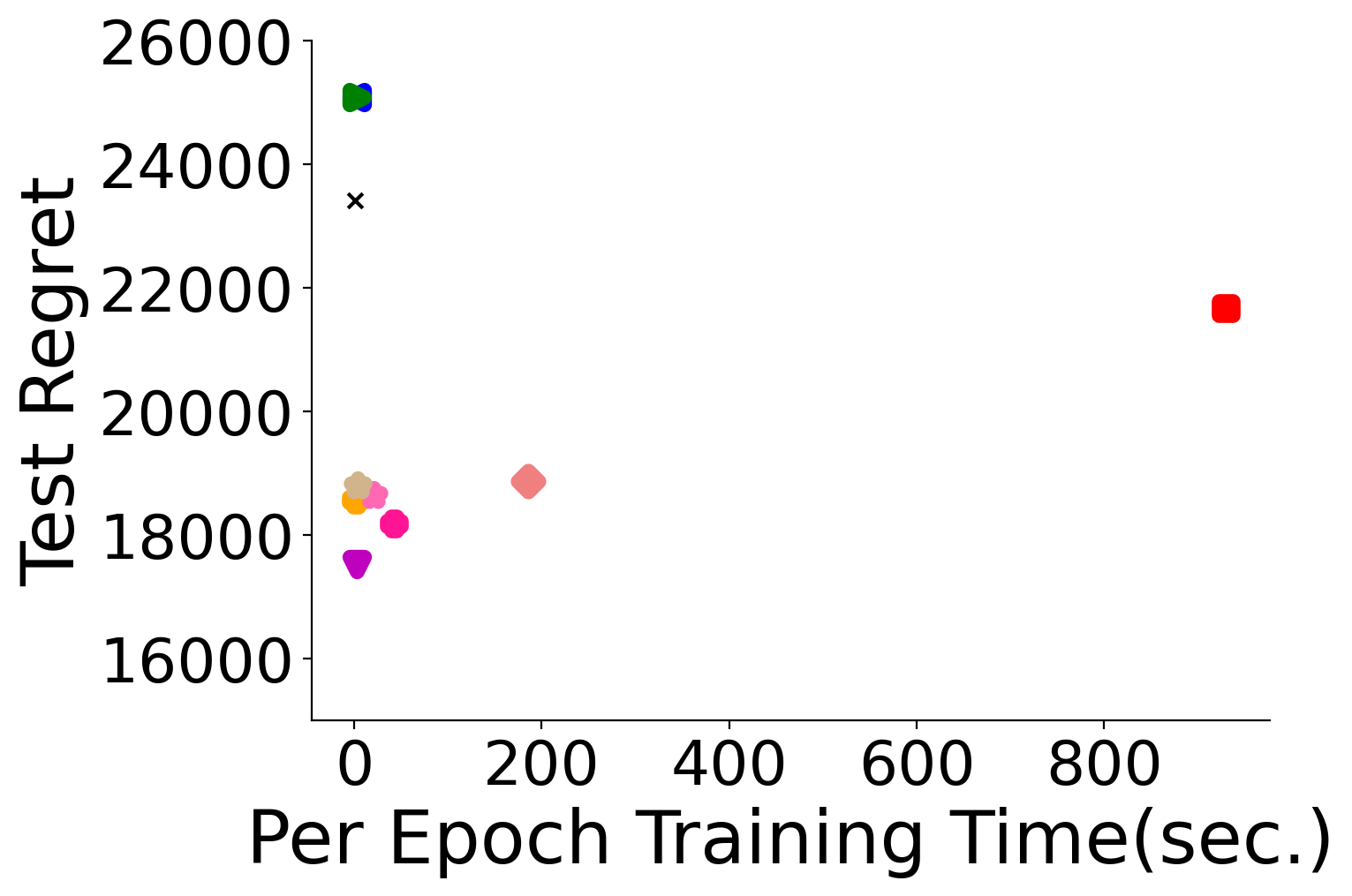}
    \caption{Energy-3}
    \label{fig:Energy}
  \end{subfigure}
  ~
  \begin{subfigure}[b]{0.28\textwidth}
    \includegraphics[width=\linewidth]{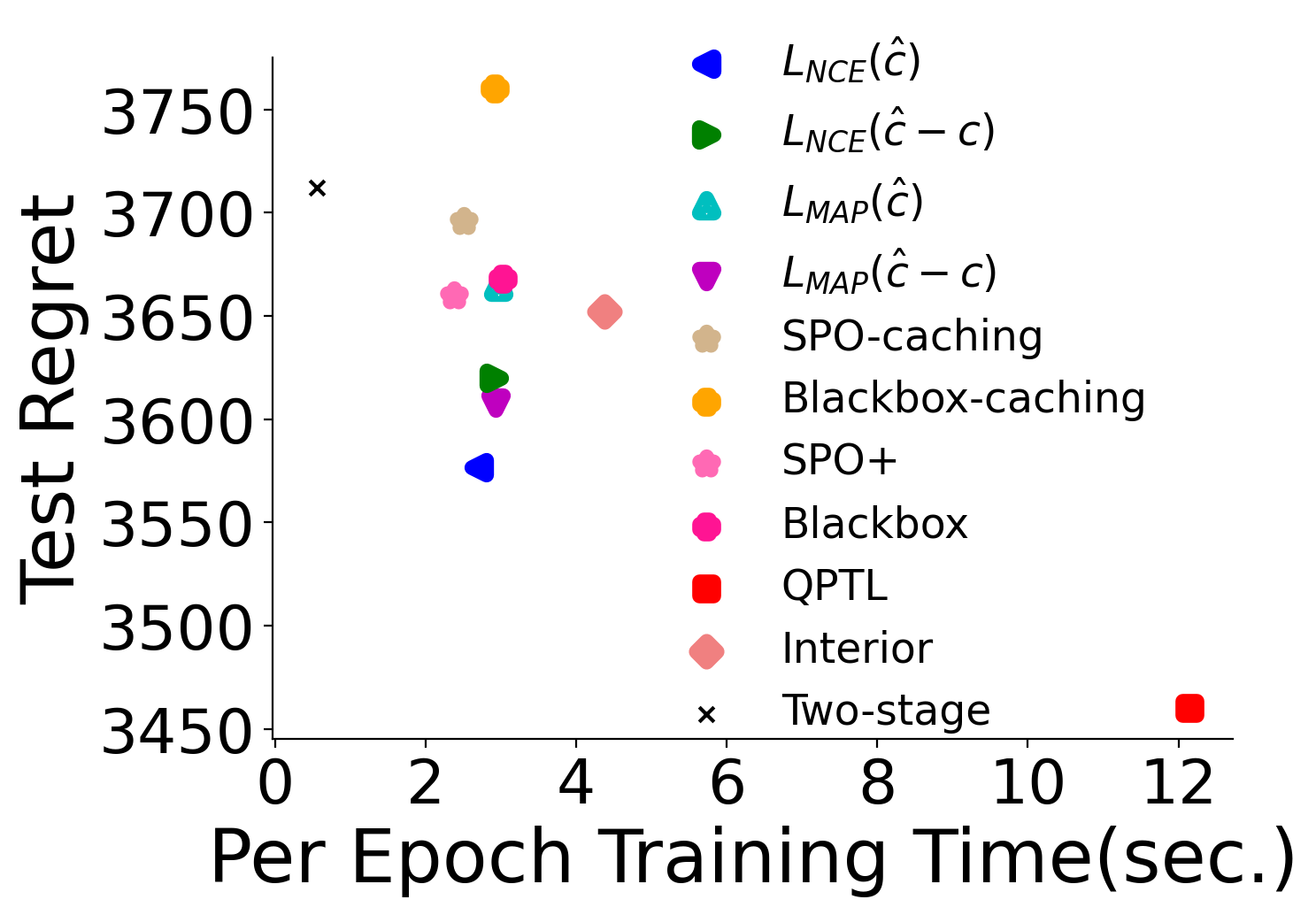}
    \caption{Matching-25}
    \label{fig:Matching}
  \end{subfigure}
  \caption{ Regret versus total training time for the different methods}
  \label{fig:q4}
\end{figure*}
To do so, we evaluate our methodology on three NP hard problems, the knapsack problem, a job scheduling problem and a maximum diverse bipartite matching problem.\footnote{ Code and data are publicly available at \url{https://github.com/CryoCardiogram/ijcai-cache-loss-pno}.}
\subsection{Experimental Settings}
\paragraph{Knapsack Problem.} The objective of this problem is to select a maximal value subset from a set of items subject to a capacity constraint.
We generate our dataset from \cite{ifrim2012properties}, which contains historical energy price data at 30-minute intervals from 2011-2013.
Each half-hour slot has features such as calendar attributes; day-ahead estimates of weather characteristics; SEMO day-ahead forecasted energy-load, wind-energy production and prices.
Each knapsack instance consists of $48$  half-our slots, which basically translates to one calendar day. 
The knapsack weights are synthetically generated where a weight $ \in \{ 3,5,7 \}$ is randomly assigned to each of the 48 slots and the price is multiplied accordingly before adding Gaussian noise $\xi \sim \mathbb{N} (0,25)$ to maintain high correlation between the prices and the weights as strongly correlated instances are difficult to solve \cite{pisinger2005hard}. 
We study three instances of this knapsack problem with capacities of 60, 120 and 180.

\paragraph{Energy-cost Aware Scheduling.}
In our next experiment, we consider a more complex combinatorial optimization problem. This combinatorial problem is taken from \emph{CSPLib}~\cite{csplib}, a library of constraint optimization problems.
In energy-cost aware scheduling \cite{csplib:prob059}, a given number of tasks, each having its own duration, power usage, resource requirement, earliest possible start and latest-possible end, must be scheduled on a certain number of machines respecting the resource capacities of the machines. A task cannot be stopped or migrated once started on a machine. The cost of energy price varies throughout the day and the goal is to find a scheduling which would minimize the total energy consumption cost. 
We use the same energy price data for this experiment.
We study three instances named Energy-1, Energy-2 and Energy-3.

\paragraph{Diverse Bipartite Matching.} We adopt this experiment from \cite{ferber2020mipaal}.
The matching instances are constructed from the CORA citation network \cite{cora2008}. The graph is partitioned into 27 sets of disjoint nodes. Diversity constraints are added to ensure there are some edges between papers of the same field as well as edges between papers of different fields.
The prediction task is to predict which edges are present using the node features. The optimization task is to find a maximum matching in the predicted graph. 
Contrary to the previous ones, here the learning task is the challenging one whereas the optimisation task is relatively simpler. 
We study three instances with varying degree of diversity constraints, Matching-10, Matching-25 and -50. 

\subsection{Results}
For all the experiments, the dataset is split on training ($70\%$), validation  ($10\%$) and test  ($20\%$) data.
The validation sets are used for selecting the best hyperparameters. The final model is run on the test data 10 times and we report the average and standard deviation (in bracket) of the 10 runs. All methods are implemented with Pytorch 1.3.1 \cite{Pytorch} and Gurobi 9.0.1 \cite{gurobi}.
%\tias{Needs exact version numbers per submission form question}
\subsubsection{Q1} In section \ref{sect-contrastive_loss}, we introduced $\mathcal{L}_{\NCE}$, $\mathcal{L}_{\MAP}$, $\mathcal{L}_{\NCE}^{(\hat{c}-c)} $ and $\mathcal{L}_{\MAP}^{(\hat{c}-c)} $. 
In Table \ref{table:Q1_knap}, \ref{table:Q1_scheduling} and \ref{table:Q1_matching}, we compare the test regret of these 4 contrastive losses. The test regret of a two-stage approach, where model training is done with no regards to the optimization task, is provided as baseline.
% namely SPO+ \citep{elmachtoub2017smart,m2019smart}, Blackbox \citep{vlastelica2019differentiation}, QPTL~\citep{wilder2019melding} and Interior~\citep{mandi2020interior}.

We can see in Table \ref{table:Q1_knap} and Table\ref{table:Q1_scheduling} for the knapsack and the scheduling problem, $\mathcal{L}_{\MAP} \ (\hat{c}-c) $ performs  the best among all the loss variants. Interestingly, with $\mathcal{L}_{\NCE}$, there is no significant advantage of the linear objective loss function ($\hat{c}-c) $); whereas in case of $\mathcal{L}_{\MAP}$, we observe significant gain by using the linear objective loss function. On the other hand, in Table \ref{table:Q1_matching}, for the matching problem $\mathcal{L}_{\NCE}$ performs slightly better than $\mathcal{L}_{\MAP}$.
\subsubsection{Q2}
In the previous experiment, the initial discrete solutions on the training data as well as all solutions obtained during training are cached to form the inner approximation of the feasible region.
But, as explained in section~\ref{sect:5}, finding the optimal $v^*(\hat{c})$ and adding it to the solution cache for all $\hat{c}$ during training is computationally expensive. Instead, now we empirically experiment with $p_{solve} = 5 \% $, i.e. where new solutions are computed only $5\%$ of the time.

In Figure~\ref{fig:NCE}, we plot regret against training time for Energy-3 (we observe similar results as shown in the appendix). There is a significant reduction in computational times as we switch to $5\%$ sampling strategy. Moreover, this does have not deleterious impact on the test regret. We conclude that adding new solutions to the solution cache by sampling seem to be an effective strategy to have good quality solutions without a high computational burden.
% Here, we are going to present the result of Energy-3, we observe similar results as shown in the appendix.
% As we expect, we observe, in Figure~\ref{fig:NCE}, 
% \subsubsection{Q2A} 
%\jay{can we make growing of SPO/BB a different RQ}
~
\subsubsection{Q3}
To investigate the validity of inner-approximation  caching  approach, we implement SPO-caching and Blackbox-caching, where we perform differentiation of SPO+ loss and Blackbox solver differentiation respectively, with $p_{solve}$ being $5\%$.
% inner approximation by solution caching for two blackbox predict-and-optimize method SPO and Blackbox, where instead of calling an oracle to solve the optimization problem, we select a solution from the solution cache by ranking which one produces better objective value. Like before, here also we grow the solution cache by a $5\%$ growing strategy. 
We again plot regret against training time in Figure~\ref{fig:SPO} and Figure~\ref{fig:BB} for SPO+ and Blackbox respectively. These figures show caching drastically reduces  training times without any significant impact on regret both for SPO+ and Blackbox differentiation.
\subsubsection{Q4} Finally we investigate what we gain by implementing $\mathcal{L}_{\NCE}$, $\mathcal{L}_{\NCE}^{(\hat{c}-c)} $, $\mathcal{L}_{\MAP}^{(\hat{c}-c)} $ and SPO-caching and blackbox-caching with $p_{solve}$ being $5\%$. We compare them against some of the state-of-the-art approaches- SPO+ \cite{elmachtoub2017smart,m2019smart}, Blackbox \cite{vlastelica2019differentiation}, QPTL~\cite{wilder2019melding} and Interior~\cite{mandi2020interior}. Our goal is not to beat them in terms of regret; rather our motivation is to reach similar regret in a time-efficient manner. 

 In Figure \ref{fig:knapsack}, Figure \ref{fig:Energy} and \ref{fig:Matching}, we plot Test regret against per epoch training time for Knapsack-120, Energy-3 and Matching-25. In Knapsack-120, Blackbox and Interior performs best in terms of regret. $\mathcal{L}_{\MAP}^{(\hat{c}-c)} $, SPO-caching and Blackbox-caching attain low regret comparable to these with a \textbf{significant gain in training time}.
 For Energy-3 the regret of SPO-caching and Blackbox-caching are comparable to the state of the art, whereas $\mathcal{L}_{\MAP}^{(\hat{c}-c)} $, in this specific case, results in lowest regret at very low training time.  
In Matching-25, QPTL is the best albeit the slowest and SPO+ and Blackbox perform marginally better than a two-stage approach. In this instance, caching methods are not good enough; but the four contrastive methods performs better than SPO+ and Blackbox. These methods can be viewed as trade-off between lower regret of QPTL and faster runtime of two-stage.
\section{Concluding Remarks}
We presented a methodology for decision-focused learning based on two main contributions: i. A new family of loss functions inspired by noise contrastive estimation; and ii. A solution cache representing an inner approximation of the feasible region. We adapted the solution caching concept to other state-of-the-art methods, namely Blackbox \cite{vlastelica2019differentiation} and SPO+ \cite{elmachtoub2017smart}, for decision-focused learning improving their efficiency. These two concepts allow to reduce solution times drastically while reaching similar quality solutions.

%We presented a methodology for decision-focused learning for optimization problems based on two main concepts: ; and i.  a new family of loss functions based on these two concepts. We adapted the solution caching concept to other state-of-the-art methods, namely Blackbox \cite{vlastelica2019differentiation} and SPO+ \cite{elmachtoub2017smart}, for decision-focused learning improving their efficiency. These two concepts allow to reduce solution times drastically while not losing quality of the solutions.

\section*{Acknowledgments}
This research received partial funding
from the Flemish Government (AI Research Program), the FWO Flanders projects G0G3220N and Data-
driven logistics (FWO-S007318N) and the H2020 Project AI4EU, G.A. 825619 as well as from the European Research Council (ERC H2020, Grant agreement No. 101002802, CHAT-Opt)
%Authors thank to Duvel :)

%\michele{notes he is more of a Westmalle guy :-)}
%\clearpage
\bibliographystyle{named}
\bibliography{ijcai21}
%\ignore{ 
\appendix

% \section{Things to put in the appendix}\label{stylefiles}

% Here will be the appendix
\newpage
\section{Predictive models}
For knapsack and scheduling experiments, we train a simple linear regressor implemented as a single layer network in PyTorch, trained to minimize Mean Square Error. For matching experiments, we consider a 2-layers perceptron with an hidden layer of 200 units and ReLU activation function. As it is used in a binary classification problem, we apply a sigmoid function on its output. The neural network is trained to minimize Cross Entropy. 

All predictive model are trained with ADAM \cite{Kingma2015AdamAM}.

\section{Detailed Result}
\begin{table*}[hbt!]
\centering
\resizebox{\linewidth}{!}{
\begin{tabular}{lcccccccccc}
\toprule
& & Two-stage & SPO & Blackbox & QPTL &  Interior  &\makecell{SPO-caching \\(5$\% $)}& \makecell{Blackbox-caching  \\(5$\% $)} \\
\midrule
\multirow{ 3}{*}{Knapsack-60}& Regret &  989 (14) & 734 (10) & 659 (15) & 668 (4) & 700 (54) & 730 (42) & 682 (37) \\
 \cline{2-9} \\
& MSE & 34101(115) & $10^{6}$ (105) & $1.5 \times 10^{6}$ (18) & $ 10^{6}$ (533) & $1.5 \times 10^{6}$ (232) &$ 10^{6}$ (8$\times 10^3$) & $1.5 \times 10^{6}$ (44)\\
\midrule
\multirow{3}{*}{Knapsack-120}& Regret & 1090 (27) & 578 (1) & 528 (10) & 545 (1) &  565 (40) & 558 (7) & 565 (40) \\
 \cline{2-9} \\
& MSE & 34107(79) & $ 10^{6}$ (50) & $1.5 \times 10^{6}$ (40) & $1.5 \times 10^{6}$ (37) & $1.5 \times 10^{6}$ (280) & $ 10^{6}$ ($10^3$) & $1.5 \times 10^{6}$ (36)\\
\midrule
\multirow{ 3}{*}{Knapsack-180}& Regret &433 (12) & 316 (1) & 314 (13) & 1461 (13) & 370 (22) & 354 (14) & 316 (22)\\
 \cline{2-9} \\
& MSE & 34145(80) & $0.5 \times 10^{6}$ (33) & $1.5 \times 10^{6}$ (16) & $1.5 \times 10^{6}$ (44) & $1.5 \times 10^{6}$ (260) & $ 10^{6}$ ($10^3$) & $1.5 \times 10^{6}$ (64)\\
\bottomrule
\end{tabular}
}
\caption{Test Regret of the state of the art for the Knapsack Problem}
\label{table:sota-Regretknapsack}
\end{table*}

\begin{table*}[ht]
\centering
% \resizebox{\linewidth}{!}{
\begin{tabular}{lcccccccc}
\toprule
 & SPO &   \makecell{SPO-caching \\(5$\% $)}& Blackbox & \makecell{Blackbox-caching  \\(5$\% $)} \\
\midrule
Knapsack-60 & 2 & 0.5 &  4.5 & 0.5 \\
\\
Knapsack-120 & 2  & 0.5 &  4 & 0.5 \\
\\
Knapsack-180 & 2 & 0.5 & 4 & 0.5\\
\bottomrule
\end{tabular}
\caption{Per Epoch Runtime (sec.) of the caching implementation for the Knapsack Problem}
\label{table:cachingtimeknapsack}
\end{table*}

% \begin{figure*}
%     \centering
%     \begin{minipage}{.33\textwidth}
%         \centering
%         \includegraphics[scale=0.35]{capa60lrcurve.png}
%         \label{fig:prob1_6_2}
%     \end{minipage}%
%     \begin{minipage}{.33\textwidth}
%         \centering
%         \includegraphics[scale=0.35]{capa120lrcurve.png}
%         \label{fig:prob1_6_2}
%     \end{minipage}%
%     \begin{minipage}{0.33\textwidth}
%         \centering
%         \includegraphics[scale=0.35]{capa180lrcurve.png}
%         % \caption{Learning Curve}
%         \label{fig:prob1_6_1}
%     \end{minipage}
%     \caption{Learning Curve for Knapsack Problem}
% \end{figure*}
\begin{table*}[ht]
\centering
\resizebox{\linewidth}{!}{
\begin{tabular}{lcccccccccccc}
\toprule
&  &\multicolumn{3}{c}{$\mathcal{L}_{\NCE}$}  & &
% \multicolumn{3}{c}{$\mathcal{L}_{ReLU}$} & &
\multicolumn{3}{c}{$\mathcal{L}_{MAP}$}  \\[0.5em]
\cline{3-5} \cline{7-9} 
% \cline{11-13} 
\\
& & \small $\hat{c}$ & \small $(\hat{c}-c) $ &  \small $(2\hat{c}-c) $ & & 
% \small $\hat{c}$ & \small $(\hat{c}-c) $ &  \small $(2\hat{c}-c) $ & &
\small $\hat{c}$ & \small $(\hat{c}-c) $ &  \small $(2\hat{c}-c) $\\
\toprule
\multirow{ 3}{*}{Knapsack-60}& Regret & 912 (21) & 1024 (66) & 908 (15) & & 
% 1942 (900) & 808 (8) & 882 (8) && 
1277 (555) & \textbf{764 (2)} & 809 (2)
\\
 \cline{2-13} \\
& MSE & $ 10^{6}$ (2$\times 10^3$)& $1.5 \times 10^{6}$ (425) & $ 10^{6}$ ($10^3$)&& 
% $1.5 \times 10^{6}$ (2)& $1.5 \times 10^{6}$ ($10^3$)& $ 10^{6}$ (2$\times10^3$) & & 
$1.5 \times 10^{6}$ (2) & $10^{6}$ (395) & $ 10^{6}$ (287)\\
\midrule
\multirow{3}{*}{Knapsack-120}& Regret & 760 (12) & 770 (15) & 763 (10) & & 
% 2177 (671) & \textbf{547 (6)}&  722 (45) & & 
912 (9) & \textbf{562 (1)} & 591 (2)\\
 \cline{2-13} \\
& MSE & $1.5 \times 10^{6}$ (3) & $1.5 \times 10^{6}$ (3) & $1.5 \times 10^{6}$ (2) && 
% $1.5 \times 10^{6}$ (2) & $10^{6}$ ($10^3$) & $ 10^{6}$  ($10^3$) & &
$1.5 \times 10^{6}$ (1) & $10^{6}$ (292) & $ 10^{6}$ (264)\\
\midrule
\multirow{ 3}{*}{Knapsack-180}& Regret & 2475 (45) & 2474 (40) & 2478 (630) && 
% 1507 (506) & 378 (3) & 365 (7) && 
491 (8) & \textbf{327 (1)} & 331 (5)
\\[0.25em]
 \cline{2-13} \\
& MSE & $ 10^{6}$ (47) & $10^{6}$ (62) & $ 10^{6}$ (24) & & 
% $1.5 \times 10^{6}$ (4) & $10^{6}$ (860) & $10^{6}$ (324) & & 
$1.5 \times 10^{6}$ (1) & 40500 (200) & 52000 (158)
& & & &\\
\bottomrule
\end{tabular}
}
\caption{Test Regret of the Variants of the Knapsack Problem}
\label{table:Regretknapsack}
\end{table*}

\begin{figure*}[ht]
  \begin{subfigure}[b]{0.32\textwidth}
    \includegraphics[width=\linewidth]{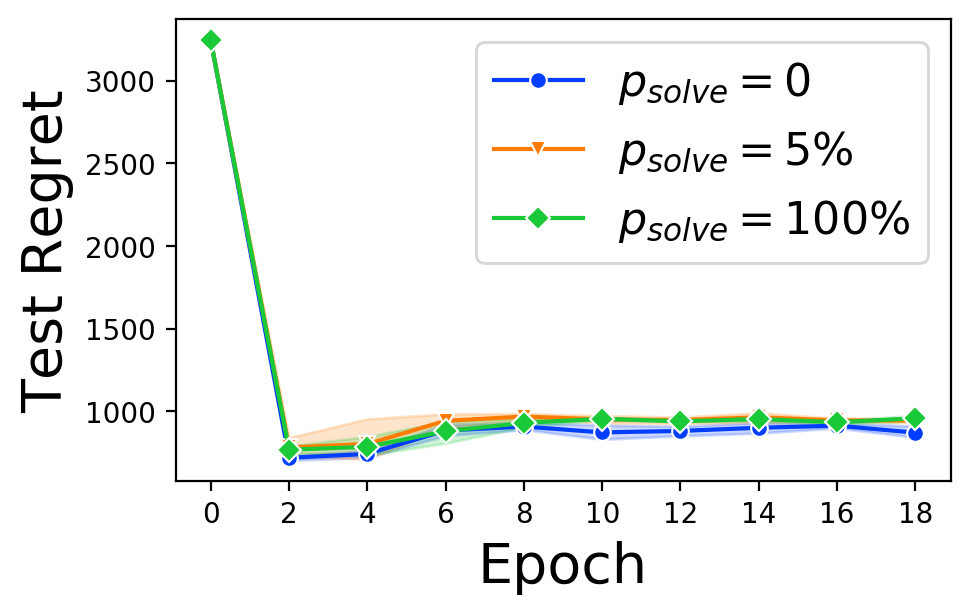}
    \caption{Knapsack-60}
    \label{fig:11}
  \end{subfigure}
  ~
  \begin{subfigure}[b]{0.32\textwidth}
    \includegraphics[width=\linewidth]{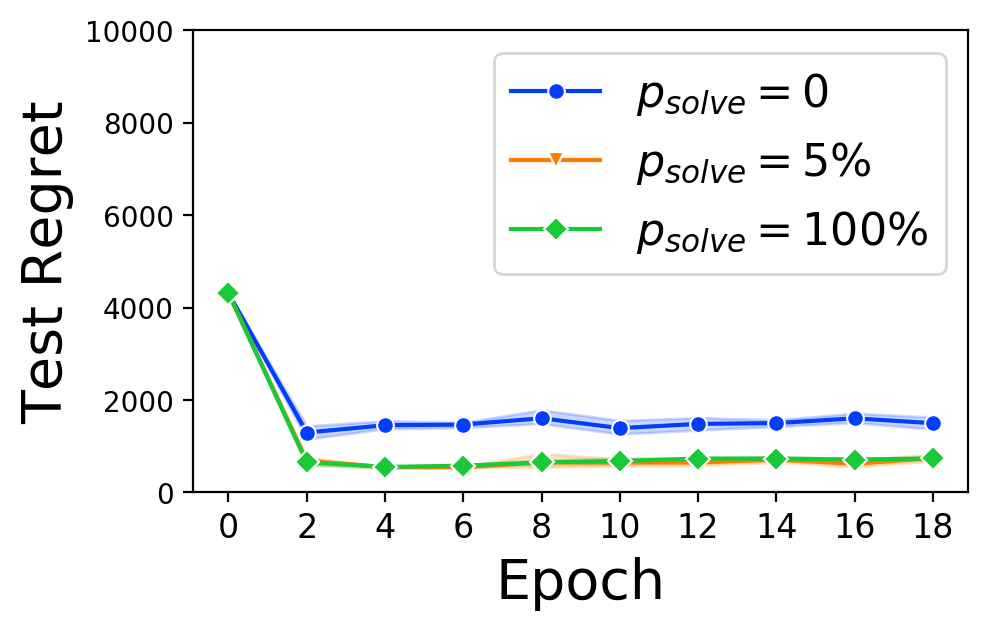}
    \caption{Knapsack-120}
    \label{fig:21}
  \end{subfigure}
  ~
  \begin{subfigure}[b]{0.32\textwidth}
    \includegraphics[width=\linewidth]{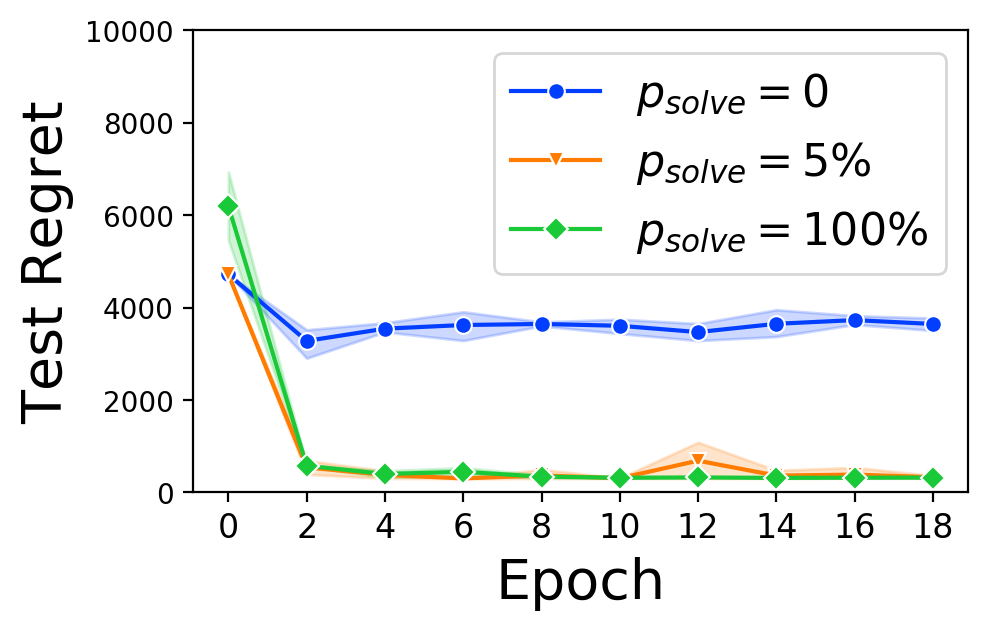}
    \caption{Knapsack-180}
    \label{fig:31}
  \end{subfigure}
    \begin{subfigure}[b]{0.32\textwidth}
    \includegraphics[width=\linewidth]{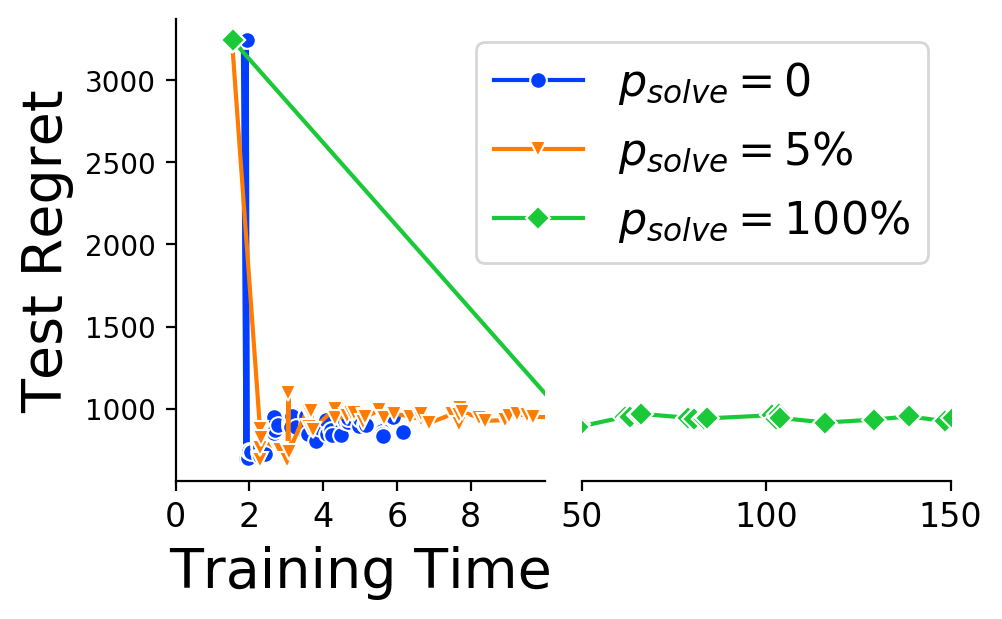}
    \caption{Knapsack-60}
    \label{fig:41}
  \end{subfigure}
  ~
  \begin{subfigure}[b]{0.32\textwidth}
    \includegraphics[width=\linewidth]{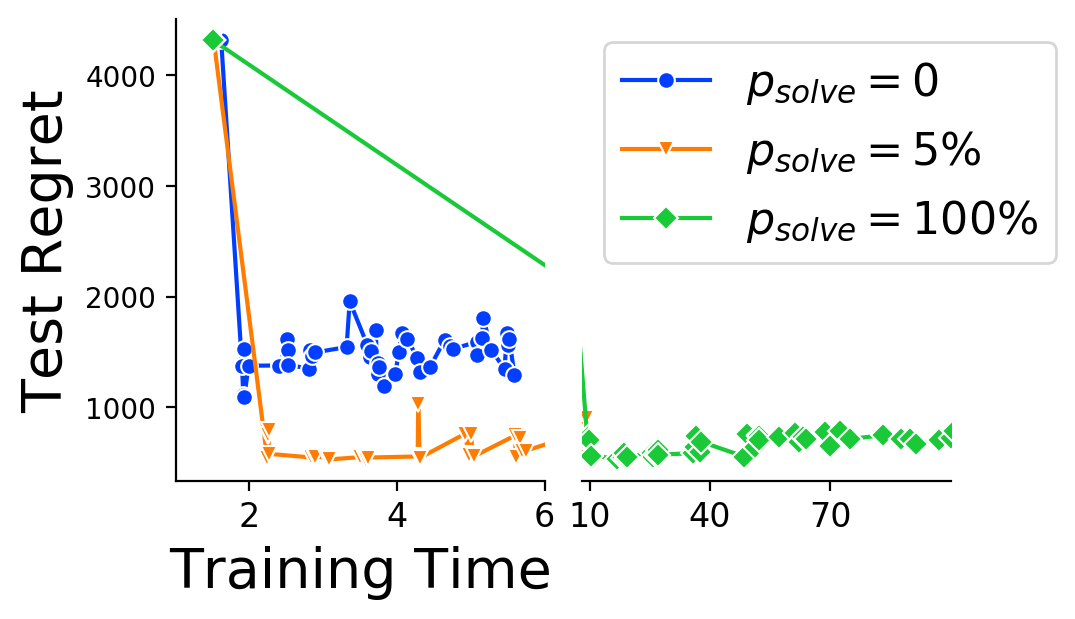}
    \caption{Knapsack-120}
    \label{fig:51}
  \end{subfigure}
  ~
  \begin{subfigure}[b]{0.32\textwidth}
    \includegraphics[width=\linewidth]{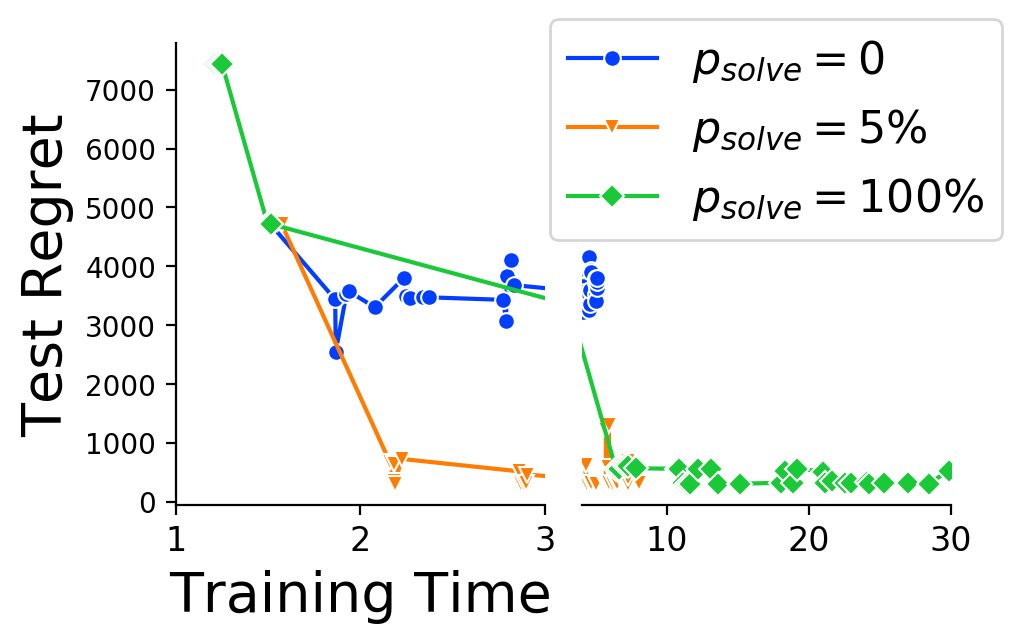}
    \caption{Knapsack-180}
    \label{fig:61}
  \end{subfigure}
  \caption{Learning Curve for Knapsack Problem \\ $\mathcal{L}_{MAP}$ $(\hat{c}-c) $ }
\end{figure*}

\begin{table*}[ht]
\centering
\resizebox{\linewidth}{!}{
\begin{tabular}{lcccccccccc}
\toprule
& & Two-stage & SPO-relax & Blackbox & QPTL & Interior  &\makecell{SPO-caching \\(5$\% $)}& \makecell{Blackbox-caching  \\(5$\% $)} \\
\midrule
\multirow{3}{*}{Energy-1 }& Regret   &43384 (376) & 40935 (115) & 38800 (79) &40615 (799) & 40200 (225)& 40820 (76)&   41635 (781) \\[0.25em]
 \cline{2-9} \\[0.05em]
& MSE & 769 (3) & 9648 (148) & 5432 (2) & 5161 (1) & 2334 (66) & 14464 (434) & 5612 (42))\\
\midrule
\multirow{ 3}{*}{Energy-2}& Regret & 31978 (781) & 27470 (42) & 27405 (144) & 37639 (194)& 29407 (1855) & 27591 (77) & 27776 (856) \\[0.25em]
 \cline{2-9} \\[0.05em]
& MSE & 768 (3) & 5582 (248) & 5525 (2) & 5436 (1) & 3651 (47)& 10654 (664) & 5571 (31)\\
\midrule
\multirow{ 3}{*}{Energy-3 }& Regret &   23423 (893)& 18647 (234) & 18187 (587) & 21677 (857) & 18509 (1121) & 18803 (39) & 18881 (1481) \\[0.25em]
 \cline{2-9} \\[0.05em]
& MSE & 770 (3) & 3237 (37) & 5485 (5) & 5208 (4) & 3742 (67)& 5008 (229) & 5455 (32)\\
\bottomrule
\end{tabular}
}
\caption{Test Regret of the state of the art of the Energy Scheduling Problem}
\label{table:sota-Regretenergy}
\end{table*}
~
\begin{table*}[ht]
\centering
% \resizebox{\linewidth}{!}{
\begin{tabular}{lcccccccc}
\toprule
 & SPO &   \makecell{SPO-caching \\(5$\% $)}& Blackbox & \makecell{Blackbox-caching  \\(5$\% $)} \\
\midrule
Energy-1 & 6.5 & 2 &  14 &  1 \\
\\
Energy-2 & 12  & 2  &  26.5 & 0.5 \\
\\
Energy-3 & 21 & 3.5 & 42.5 & 1.5\\
\bottomrule
\end{tabular}
\caption{Per Epoch Runtime (sec.) of the caching implementation for the Energy Scheduling Problem}
\label{table:cachingtimeenergy}
\end{table*}
~
\begin{figure*}[ht]
  \begin{subfigure}[b]{0.32\textwidth}
    \includegraphics[width=\linewidth]{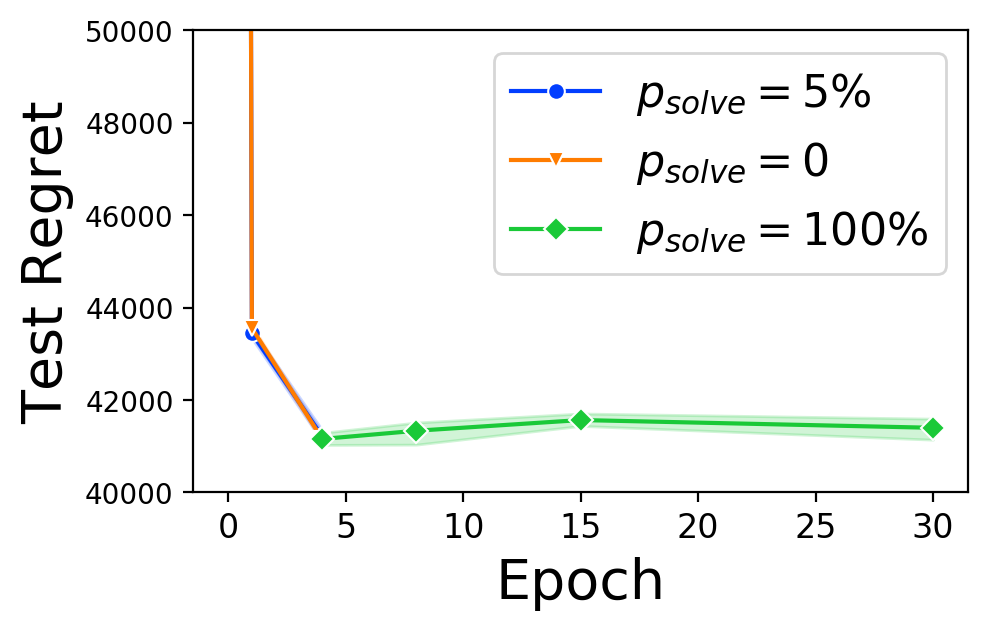}
    % \caption{}
    \label{fig:1}
  \end{subfigure}
  ~
  \begin{subfigure}[b]{0.32\textwidth}
    \includegraphics[width=\linewidth]{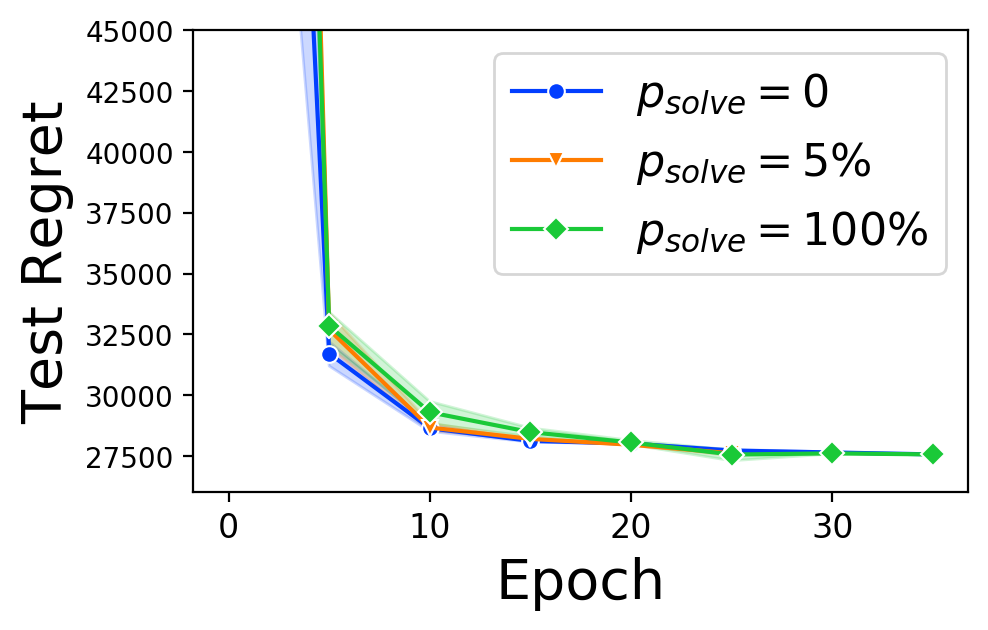}
    % \caption{}
    \label{fig:2}
  \end{subfigure}
  ~
  \begin{subfigure}[b]{0.32\textwidth}
    \includegraphics[width=\linewidth]{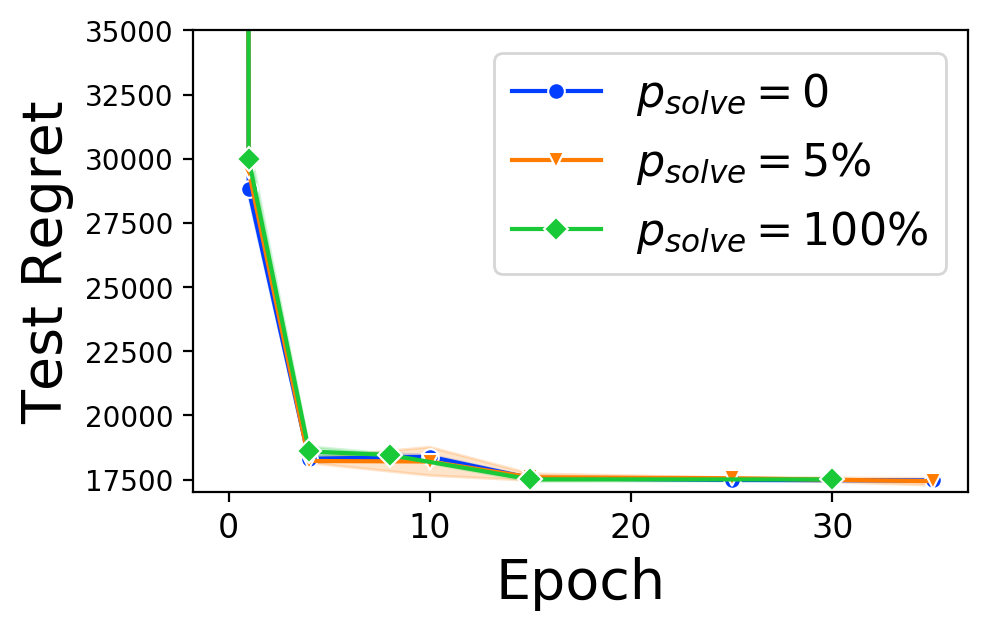}
    % \caption{}
    \label{fig:3}
  \end{subfigure}
  \begin{subfigure}[b]{0.32\textwidth}
    \includegraphics[width=\linewidth]{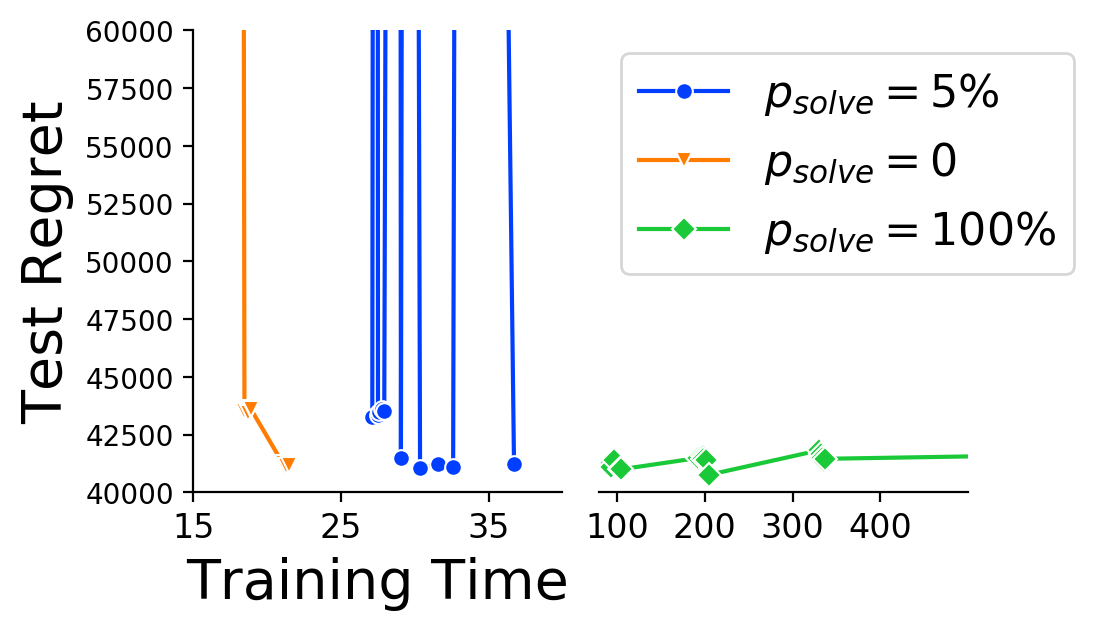}
    \caption{Energy-1}
    \label{fig:4}
  \end{subfigure}
  ~
  \begin{subfigure}[b]{0.32\textwidth}
    \includegraphics[width=\linewidth]{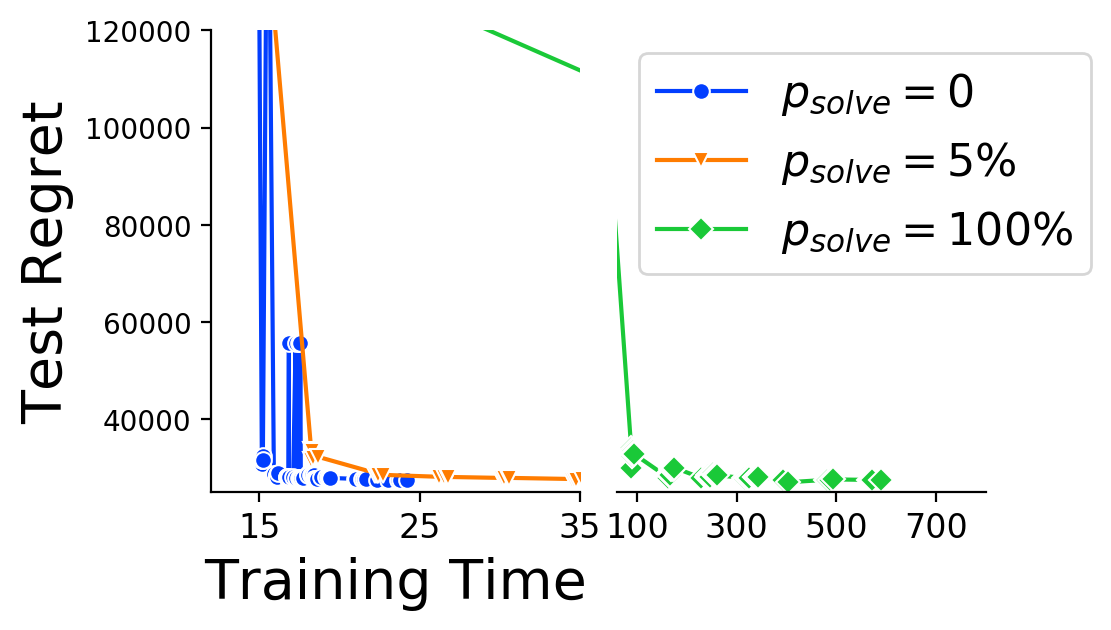}
    \caption{Energy-2}
    \label{fig:5}
  \end{subfigure}
  ~
  \begin{subfigure}[b]{0.32\textwidth}
    \includegraphics[width=\linewidth]{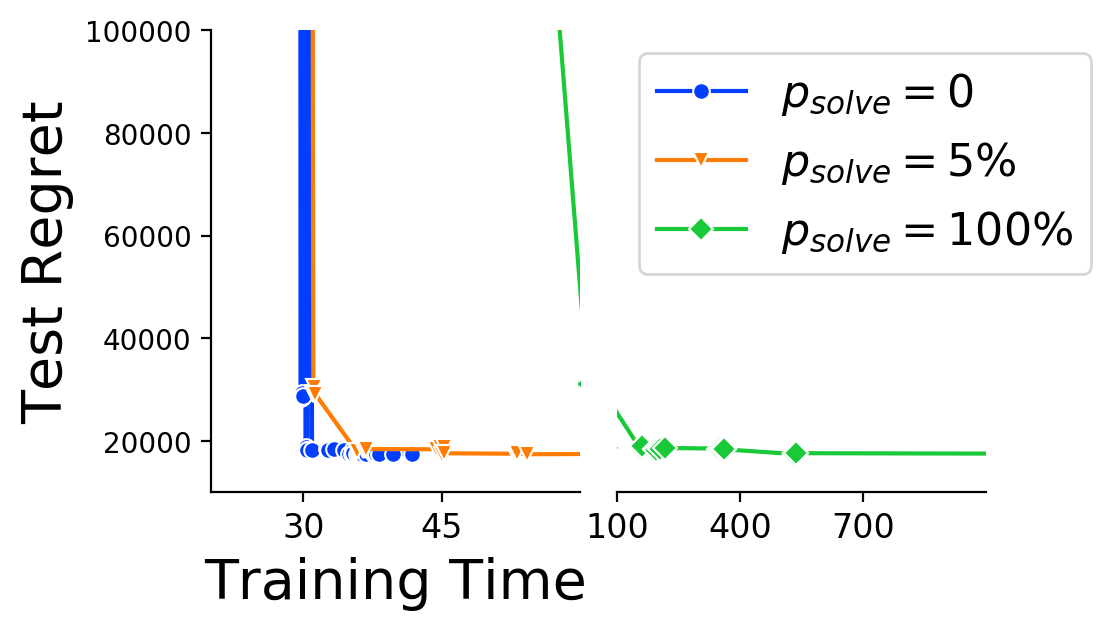}
    \caption{Energy-3}
    \label{fig:6}
  \end{subfigure}
  \caption{Learning Curve for Energy Scheduling Problem \\ $\mathcal{L}_{MAP}$ $(\hat{c}-c) $ }
\end{figure*}
\begin{table*}[ht]
\centering
\resizebox{\linewidth}{!}{
\begin{tabular}{lccccccccccc}
\toprule
&  & \multicolumn{3}{c}{$\mathcal{L}_{\NCE}$}  &
% \multicolumn{3}{c}{$\mathcal{L}_{ReLU}$} &
\multicolumn{3}{c}{$\mathcal{L}_{MAP}$}  \\[0.5em]
\cline{4-4} \cline{7-7} \\
& &\small $\hat{c}$ & \small $(\hat{c}-c) $ &  \small $(2\hat{c}-c) $ &
% \small $\hat{c}$ & \small $(\hat{c}-c) $ &  \small $(2\hat{c}-c) $ &
\small $\hat{c}$ & \small $(\hat{c}-c) $ &  \small $(2\hat{c}-c) $\\
% \begin{small}
\toprule
 \multirow{ 3}{*}{Energy-1 }& Regret & 45847(780) & 45834(1657) & 51504(339) &  
%  97341(2642) & 41955(91) & 45017(3798) & 
 104496 (18109) & 41236 (66) & \textbf{40762 (85)}\\[0.25em]
 \cline{2-11} \\[0.05em]
& MSE & 5477(5) & 5475(4) & 5352(268) & 
% 6352(36) & 6045(47) & 5705(344) & 
19070(291) & 12958(1753) & 15133(1164)\\
\midrule
 \multirow{ 3}{*}{Energy-2 }& Regret & \textbf{27633(214)} & 28994(659) & 29339(681) &  
%  43270 (16486) & 29834(34)&  29603(142) &  
 50897(20958) & 27734 (267) &  27728(346)\\[0.25em]
 \cline{2-11} \\[0.05em]
& MSE & 5578(55) & 17146(2620) & 17864(3613) & 
% 6501(506) & 5189(103) & 5441(500) & 
6190(14) & 10528(945) & 11580(858)\\
\midrule
\multirow{ 3}{*}{Energy-3}& Regret & 18789(194) & 18768(406) & 18665(630) & 
% 69340(60895) & 22050(219) & 25423(581) & 
32180(8382) & \textbf{17507(42)} & 17504(156)\\[0.25em]
 \cline{2-11} \\[0.05em]
 & MSE & 14528(578) & 14814(742) & 13691(1572) & 
%  3993(238) & 4794(353) & 3925(286) & 
 5876(36) & 33593(4992) & 49130(13108) & & &\\
\bottomrule
% \end{small}
\end{tabular}
}
\caption{Test Regret of the Variants of the Energy Scheduling Problem}
\label{table:Regretenergy}
\end{table*}

\begin{table*}[hbt!]
\centering
\resizebox{\linewidth}{!}{
	\begin{tabular}{lcccccccccc}
		\toprule
		& & Two-stage & SPO & Blackbox & QPTL &  Interior  &\makecell{SPO-caching \\(5$\% $)}& \makecell{Blackbox-caching  \\(5$\% $)} \\
		\midrule
		\multirow{ 3}{*}{Matching-10}& Regret &  3700 (42) & 3746	(54) & 3642	(73)& 3422	(80) & 3788 (58) & 3664	(140) &  3704	(74) \\
		\cline{2-9} \\
		& AUC & 0.497	(0.001) & 0.492	(0.010) & 0.483	(0.003) & 0.586	(0.019) & 0.495	(0.015) &0.480	(0.001) & 0.495	(0.005)\\
		\midrule
		\multirow{3}{*}{Matching-25}& Regret & 3712	(59) & 3786	(85) & 3710	(65) & 3482	(71) & 3652	(87) & 3696	(85) & 3760	(57) \\
		\cline{2-9} \\
		& AUC & 0.495	(0.001) & 0.494	(0.014) & 0.503	(0.006) & 0.569	(0.016) & 0.501	(0.003) & 0.498	(0.004) & 0.498	(0.013)\\
		\midrule
		\multirow{ 3}{*}{Matching-50}& Regret &3440	(36) & 3458	(61) & 3314	(76) & 3242	(75) & 3524	(117)	 & 3416	(70) & 3440	(28)\\
		\cline{2-9} \\
		& AUC & 0.494	(0.003) & 0.510	(0.012) & 0.525	(0.004) & 0.564	(0.011) & 0.485	(0.002) & 0.507	(0.003) & 0.490	(0.008)\\
		\bottomrule
	\end{tabular}
}
\caption{Test Regret of the state of the art for the Diverse Bipartite Matching Problem}
\label{table:sota-Regretmatching}
\end{table*}

\begin{table*}[h]
\centering
\resizebox{\linewidth}{!}{
	\begin{tabular}{lcccccccccccc}
		\toprule
		&  &\multicolumn{3}{c}{$\mathcal{L}_{\NCE}$}  & & 
% 		\multicolumn{3}{c}{$\mathcal{L}_{ReLU}$} & & 
		\multicolumn{3}{c}{$\mathcal{L}_{MAP}$}  \\[0.5em]
		\cline{3-5} \cline{7-9} \cline{11-13} \\
		& & \small $\hat{c}$ & \small $(\hat{c}-c) $ &  \small $(2\hat{c}-c) $ & & 
% 		\small $\hat{c}$ & \small $(\hat{c}-c) $ &  \small $(2\hat{c}-c) $ & & 
		\small $\hat{c}$ & \small $(\hat{c}-c) $ &  \small $(2\hat{c}-c) $\\
		\toprule
		\multirow{ 3}{*}{Matching-10}& Regret & 3702	(64) & 3618	(81) & 3710	(45) & & 
% 		3696	(74) &3636	(98) & 3664	(63)	 && 
		3708	(88) & 3732	(85) & 3692	(63)
		\\
		\cline{2-13} \\
		& AUC & 0.486	(0.003)& 0.487	(0.004) & 0.488	(0.007)&& 
% 		0.493	(0.003)& 0.483	(0.003)& 0.485	(0.004) & & 
		0.495	(0.002) & 0.493	(0.004) & 0.492	(0.002)\\
		\midrule
		\multirow{3}{*}{Matching-25}& Regret & 3696	(76) & 3674	(48) & 3736	(74) & & 
% 		3734	(54)	 & 3690	(67)&  3698	(67) & & 
		3700	(23)	 &3712	(86) & 3760	(60)\\
		\cline{2-13} \\
		& MSE & 0.496	(0.004) & 0.495	(0.002) & 0.496	(0.006) && 
% 		0.495	(0.003) & 0.494	(0.004) & 0.492	(0.003) & & 
		0.494	(0.002) & 0.490	(0.005) &0.494	(0.004)\\
		\midrule
		\multirow{ 3}{*}{Matching-50}& Regret & 3382 (49)&	3376 (73)&	3396 (57)	 && 
% 		3418 (77)&	3374 (69)&	3322 (73) && 
		3444 (74)&	3402 (66)&	3380 (86)	
		\\[0.25em]
		\cline{2-13} \\
		& AUC & 0.513 (0.004)&	0.517 (0.005)&	0.512 (0.003)	 & & 
% 		0.5 (0.007)&	0.529 (0.008)&	0.522 (0.006)	 & & 
		0.499 (0.002)&	0.509 (0.01)&	0.506 (0.007)
		& & & \\
		\bottomrule
	\end{tabular}
}
\caption{Test Regret of the Variants of the Diverse Bipartite Matching Problem}
\label{table:Regretmatching}
\end{table*}

\section{Problem Specification}
\subsection{Knapsack problem}
In this problem we have to predict electricity-price $c_t$ of $T$(= 48) half-hour slots of the next day. Each slot $t$, is associated with a known weight $w_t$, which can be interpreted as the agency fee of that slot. The problem of maximizing the total revenue is stated as the following MILP:

\begin{align*}
    \max_x &\sum_{t \in T} c_t x_t& \\
    \text{subject to }& \sum_{t \in T}w_t x_t \leq B & \\
    &x_t \in \{0,1\}& t \in T
\end{align*}

where $x_t$ is a binary variable which takes the value 1 only if slot $t$ is purchased.  Constraint limits the number of slots to buy under a fixed budget $B$. The objective is to make maximum revenue by selling the electricity to the grid at price $c_t$ in the slots already purchased. 

\subsection{Energy-cost Aware Scheduling}
In this problem $J$ is the set of tasks to be scheduled on $M$ number of machines maintaining resource requirement of $R$ resources. The tasks must be scheduled over $T$(= 48) set of equal length time periods.
Each task $j$ is specified by its duration $d_j$, earliest start time $e_j$, latest end time $l_j$, power usage $p_j$.$u_{jr}$ is the resource usage of task $j$ for resource $r$ and $c_{mr}$ is the capacity of machine $m$ for resource $r$.
Let $x_{jmt}$ be a binary variable which is 1 only if task $j$ starts at time $t$ on machine $m$.
If $c_t$ is the (predicted) energy price at timeslot $t$, the objective is to minimize the following formulation of total energy cost.
Then the problem can be formulated as  the following MILP:
\begin{align*}
\min_{x_{jmt}} & \sum_{j \in J} \sum_{m \in M} \sum_{t \in T} x_{jmt} \Big( \sum_{t \leq t' < t+d_j} p_j c_{t'} \Big)\\
\text{subject to }& \sum_{m \in M} \sum_{t \in T} x_{jmt} =1 \ , \forall_{j \in J}\\
& x_{jmt} = 0 \ \ \forall_{j \in J} \forall_{m \in M} \forall_{t < e_j}\\
& x_{jmt} = 0 \ \ \forall_{j \in J} \forall_{m \in M} \forall_{t + d_j > l_j} \\
& \sum_{j \in J} \sum_{t - d_{j} < t' \leq t} x_{jmt' } u_{jr}  \leq c_{mr},  \forall_{m \in M} \forall_{r \in R} \forall_{t \in T}
\end{align*}
The first constraint ensures each task is scheduled only once.
% \[
% \sum_{m \in M} \sum_{t \in T} x_{jmt} =1 \ , \forall_{j \in J}
% \]
The next constraints ensure the task scheduling abides by earliest start time and latest end time constraints.
% \[
% x_{jmt} = 0 \ \ \forall_{j \in J} \forall_{m \in M} \forall_{t < e_j}
% \]
% \[
% x_{jmt} = 0 \ \ \forall_{j \in J} \forall_{m \in M} \forall_{t + d_j > l_j}
% \]
The final constraint is to respect the resource requirement of the machines.
% \[
% \sum_{j \in J} \sum_{t - d_{j} < t' \leq t} x_{jmt' } u_{jr}  \leq c_{mr},  \forall_{m \in M} \forall_{r \in R} \forall_{t \in T}
% \]
% \[
% \min_{x_{jmt}} \sum_{j \in J} \sum_{m \in M} \sum_{t \in T} x_{jmt} \Big( \sum_{t \leq t' < t+d_j} p_j c_{t'} \Big)
% \]
\subsection{Diverse Bipartite Matching}
Following \cite{ferber2020mipaal}, in this problem 100 nodes representing scientific publications are split into two sets of 50 nodes $N_1$ and $N_2$. A (predicted) reward matrix $c$  indicates the likelihood that a link between each pair of nodes from $N_1$ to $N_2$ exists. Finally a indicator $m_{i,j}$ is $0$ if article $i$ and $j$ share the same subject field, and $0$ otherwise $\forall i \in N_1, j \in N_2$.
Let $p \in [0,1]$ be the rate of pair sharing their field in a solution and $q \in [0,1]$ the rate of unrelated pairs, the goal is to match as much nodes in $N_1$ with at most one node in $N_2$ by selecting edges which maximizes the sum of rewards under similarity and diversity constraints. Formally:
\begin{equation*}
\begin{array}{lll}
\max_x & \sum_{i, j} c_{i, j} x_{i, j} & \\
\text { subject to } & \sum_{j} x_{i, j} \leq 1 & \forall i \in N_{1} \\
& \sum_{i} x_{i, j} \leq 1 & \forall j \in N_{2} \\ 
& \sum_{i . j} m_{i, j} x_{i, j} \geq p \sum_{i, j} x_{i, j} & \\
& \sum_{i . j} (1-m_{i, j}) x_{i, j} \geq q \sum_{i, j} x_{i, j} \\ 
& x_{i,j} \in \{0,1\} & \forall i \in N_1, j \in N_2 \\
\end{array}
\end{equation*}
In our experiments, we considered three variation of this problem with $p=q=$ $10\%$, $25\%$ and $5\%$, respectively named Matching-10, Matching-25 and Matching-50.
\section{Hyperparameter Details}
\begin{table*}[ht]

\centering
% \resizebox{\linewidth}{!}{
\begin{tabular}{lcccccccccc}
\toprule
& & Knapsack-60 & Knapsack-120  & Knapsack-180 \\
 \midrule
\multirow{ 2}{*}{Two-stage } & Learning rate &  $0.1$ & $0.1$ & $0.1$\\[0.5em]
 & Epochs & 20 & 20 & 20\\
\midrule
\multirow{ 2}{*}{SPO } & Learning rate &  $0.7$ & $0.7$ & $0.7$\\[0.5em]
 & Epochs & 4 & 20 & 20\\
 \midrule
\multirow{ 3}{*}{Blackbox }  & Learning rate & 0.01 & 0.01 & 0.01 &\\[0.5em]
& \makecell{Displace parameter\\Lambda} & 10$^ {-5}$ & 10$^ {-5}$ & 10$^ {-5}$\\[1em]
& Epochs & 32 & 32 & 32\\
\midrule
\multirow{ 3}{*}{QPTL}  & Learning rate & 0.1 & 0.01 & 0.01 &\\[0.5em]
& \makecell{Quadratic regularizer \\parameter } & 10$^ {-5}$ & 10$^ {-5}$ & 10$^ {-5}$\\[1em]
& Epochs & 24 & 24 & 24\\
\midrule
\multirow{ 4}{*}{Interior}  & Learning rate & 0.1 & 0.1 & 0.1 &\\[0.5em]
& $\lambda$cut-off & 10$^ {-3}$ & 10$^ {-4}$ & 10$^ {-3}$\\[0.5em]
& damping factor & 0.1& 0.1 & 10$^ {-3}$\\[0.5em]
& Epochs & 5 & 25 & 20\\
\midrule
\multirow{ 2}{*}{ \makecell{SPO-caching \\ 5$\% $} } & Learning rate &  $0.7$ & $0.7$ & $0.7$\\[0.5em]
& Epochs & 8 & 8 & 8\\
\midrule 
\multirow{ 3}{*}{ \makecell{Blackbox-caching  \\(5$\% $)} } & Learning rate & 0.01 & 0.01 & 0.01 &\\[0.5em]
& \makecell{Displace parameter\\Lambda} & 10$^ {-5}$ & 10$^ {-5}$ & 10$^ {-5}$\\[1em]
& Epochs & 32 & 32 & 32\\
\midrule 
\multirow{ 2}{*}{$\mathcal{L}_{\NCE}$ ($ \hat{c}$) }  & Learning rate & 0.1 & 0.001 & 0.01 \\[0.5em]
& Epochs & 20 & 20 & 20\\
\midrule 
\multirow{ 2}{*}{$\mathcal{L}_{\NCE}$ ($ \hat{c}-c $) }  & Learning rate & 0.01 & 0.001 & 0.01 \\[0.5em]
& Epochs & 20 & 20 & 20\\
\midrule 
\multirow{ 2}{*}{$\mathcal{L}_{\NCE}$ ($2\hat{c}-c $) }  & Learning rate & 0.1 & 0.001 & 0.01\\[0.5em]
& Epochs & 20 & 20 & 20\\
\midrule 
% \multirow{ 2}{*}{$\mathcal{L}_{ReLU}$ ($ \hat{c}$) }  & Learning rate & 0.01 & 0.01 & 0.01\\[0.5em]
% & Epochs & 20 & 20 & 20\\
% \midrule 
% \multirow{ 2}{*}{$\mathcal{L}_{ReLU}$ ($ \hat{c}-c $) }  & Learning rate & 0.7 & 0.7 & 0.7\\[0.5em]
% & Epochs & 20 & 20 & 20\\
% \midrule 
% \multirow{ 2}{*}{$\mathcal{L}_{ReLU}$ ($2\hat{c}-c $) }  & Learning rate & 0.7 & 0.7 & 0.7\\[0.5em]
% & Epochs & 20 & 20 & 20\\
% \midrule 
\multirow{ 2}{*}{$\mathcal{L}_{MAP}$ ($ \hat{c}$) }  & Learning rate & 0.01 & 0.001 & 0.001\\[0.5em]
& Epochs & 20 & 20 & 20\\
\midrule 
\multirow{ 2}{*}{$\mathcal{L}_{MAP}$ ($ \hat{c}-c $) }  & Learning rate & 0.7 & 0.7 & 0.7\\[0.5em]
& Epochs & 20 & 20 & 20\\
\midrule 
\multirow{ 2}{*}{$\mathcal{L}_{MAP}$ ($2\hat{c}-c $) }  & Learning rate & 0.7 & 0.7& 0.7\\[0.5em]
& Epochs &  20 & 20 & 20\\
\bottomrule
\end{tabular}
\caption{Choice of Hyperparameters (Knapsack Problem)}
% \label{table:hyperparams}
\end{table*}
\begin{table*}[!htbp]

\centering
% \resizebox{\linewidth}{!}{
\begin{tabular}{lcccccccccc}
\toprule
& & Energy-1 & Energy-2  & Energy-3 \\
\midrule
\multirow{ 2}{*}{Two-stage } & Learning rate &  $0.1$ & $0.1$ & $0.1$\\[0.5em]
 & Epochs & 20 & 20 & 20\\
 \midrule
\multirow{ 2}{*}{SPO } & Learning rate &  $0.1$ & $0.1$ & $0.1$\\[0.5em]
 & Epochs & 16 & 16 & 16\\
  \midrule
\multirow{ 3}{*}{Blackbox }  & Learning rate & 10$^ {-3}$ & 10$^ {-3}$ & 10$^ {-3}$&\\[0.5em]
& \makecell{Displace parameter\\Lambda} & 10$^ {-3}$ & 10$^ {-3}$ & 10$^ {-3}$\\[1em]
& Epochs & 30 & 30 & 30\\
\midrule
\multirow{ 3}{*}{QPTL}  & Learning rate & 0.1 & 10$^ {-3}$  & 10$^ {-2}$  &\\[0.5em]
& \makecell{Quadratic regularizer \\parameter }& 10$^ {-6}$ & 0.1 & 10$^ {-6}$ \\[1em]
& Epochs &  5& 5 & 5\\
\midrule
\multirow{ 4}{*}{Interior}  & Learning rate & 0.7 & 0.7 & 0.7 &\\[0.5em]
& $\lambda$cut-off & 0.1 & 0.1 & 0.1\\[0.5em]
& damping factor & 10$^ {-6}$& 10$^ {-6}$ & 10$^ {-6}$\\[0.5em]
& Epochs & 15 & 4 & 4\\
\midrule
\multirow{ 2}{*}{ \makecell{SPO-caching \\ (5$\%) $} } & Learning rate &  $0.1$ & $0.1$ & $0.1$\\[0.5em]
& Epochs & 20 & 20 & 20\\
\midrule 
\multirow{ 3}{*}{ \makecell{Blackbox-caching  \\(5$\% $)} } & Learning rate & 0.01 & 0.01 & 0.01 &\\[0.5em]
& \makecell{Displace parameter\\Lambda} & 10$^ {-3}$ & 10$^ {-3}$ & 10$^ {-3}$\\[1em]
& Epochs & 30 & 30 & 30\\
\midrule 
\multirow{ 2}{*}{$\mathcal{L}_{\NCE}$ ($ \hat{c}$) }  & Learning rate & 10$^ {-3}$ & 0.01 & 0.1 \\[0.5em]
& Epochs & 24 & 24 & 24\\
\midrule 
\multirow{ 2}{*}{$\mathcal{L}_{\NCE}$ ($ \hat{c}-c $) }  & Learning rate & 10$^ {-3}$ & 0.1 & 0.1 \\[0.5em]
& Epochs & 24 & 24 & 24\\
\midrule 
\multirow{ 2}{*}{$\mathcal{L}_{\NCE}$ ($2\hat{c}-c $) }  & Learning rate & 10$^ {-3}$ & 0.1 & 0.1\\[0.5em]
& Epochs & 24 & 24 & 24\\
\midrule 
% \multirow{ 2}{*}{$\mathcal{L}_{ReLU}$ ($ \hat{c}$) }  & Learning rate & 0.01 & 0.1 & 0.1\\[0.5em]
% & Epochs & 24 & 24 & 24\\
% \midrule 
% \multirow{ 2}{*}{$\mathcal{L}_{ReLU}$ ($ \hat{c}-c $) }  & Learning rate & 0.01 & 0.01 & 0.1\\[0.5em]
% & Epochs & 24 & 24 & 24\\
% \midrule 
% \multirow{ 2}{*}{$\mathcal{L}_{ReLU}$ ($2\hat{c}-c $) }  & Learning rate & 0.01 & 0.1 & 0.1\\[0.5em]
% & Epochs & 24 & 24 & 24\\
% \midrule 
\multirow{ 2}{*}{$\mathcal{L}_{MAP}$ ($ \hat{c}$) }  & Learning rate & 0.1 & 0.01 & 0.01\\[0.5em]
& Epochs & 24 & 24 & 24\\
\midrule 
\multirow{ 2}{*}{$\mathcal{L}_{MAP}$ ($ \hat{c}-c $) }  & Learning rate & 0.1 & 0.1 & 0.7\\[0.5em]
& Epochs & 24 & 24 & 24\\
\midrule 
\multirow{ 2}{*}{$\mathcal{L}_{MAP}$ ($2\hat{c}-c $) }  & Learning rate & 0.1 & 0.1 & 0.7\\[0.5em]
& Epochs &  24 & 24 & 24\\
\bottomrule
\end{tabular}
\caption{Choice of Hyperparameters (Energy Scheduling Problem) }
% \label{table:hyperparams}
\end{table*}
\begin{table*}[!htbp]
	
	\centering
	% \resizebox{\linewidth}{!}{
	\begin{tabular}{lcccccccccc}
		\toprule
		& & Matching-10 & Matching-25  & Matching-50 \\
		\midrule
		\multirow{ 2}{*}{Two-stage} & Learning rate &  $10^{-3}$ & $10^{-3}$ & $10^{-3}$\\[0.5em]
		& Epochs & 2 & 5 & 2\\
		\midrule
		\multirow{ 2}{*}{SPO } & Learning rate &  10$^{-3}$ & 10$^{-3}$ & 5$\times$ 10$^{-3}$\\[0.5em]
		& Epochs & 5 & 3 & 4\\
		\midrule
		\multirow{ 3}{*}{Blackbox }  & Learning rate & 5 $\times$10$^ {-3}$ & 5 $\times$10$^ {-3}$ & 10$^ {-2}$&\\[0.5em]
		& \makecell{Displace parameter\\Lambda} & $0.1$ & 10$^{-5}$ & $0.1$\\[1em]
		& Epochs & 3 & 5 & 4\\
		\midrule
        \multirow{ 3}{*}{QPTL}  & Learning rate & 10$^ {-2}$ & 10$^ {-2}$  & 10$^ {-3}$  &\\[0.5em]
        & \makecell{Quadratic regularizer \\parameter }& 0.1 & 10$^ {-2}$ & 10$^ {-4}$ \\[1em]
        & Epochs &  4& 8 & 7\\
        \midrule
        \multirow{ 4}{*}{Interior}  & Learning rate & 10$^ {-3}$ & 10$^ {-3}$ & 0.1 \\[0.5em]
        & $\lambda$cut-off & 10$^ {-6}$ & 0.1 & 10$^ {-6}$\\[0.5em]
        & damping factor & 5 $\times$10$^ {-2}$& 5 $\times$10$^ {-2}$ & 5 $\times$10$^ {-2}$\\[0.5em]
        & Epochs & 3 & 7 & 1\\
        \midrule
		\multirow{ 2}{*}{ \makecell{SPO-caching \\ (5$\%) $} } & Learning rate & 10$^ {-3}$  & 10$^ {-3}$  & 5 $\times$10$^ {-3}$\\[0.5em]
		& Epochs & 3 & 4 & 6\\
		\midrule 
		\multirow{ 3}{*}{ \makecell{Blackbox-caching  \\(5$\% $)} } & Learning rate &  10$^ {-3}$  & 10$^ {-3}$  & 5 $\times$10$^ {-3}$\\[0.5em]
		& \makecell{Displace parameter\\Lambda} & 20 & 20 & 10 \\[1em]
		& Epochs & 4 & 4 & 2 \\
		\midrule 
		\multirow{ 2}{*}{$\mathcal{L}_{\NCE}$ ($ \hat{c}$) }  & Learning rate & 10$^ {-3}$ & 10$^ {-3}$ & 10$^ {-3}$ \\[0.5em]
		& Epochs & 4 & 4 & 3\\
		\midrule 
		\multirow{ 2}{*}{$\mathcal{L}_{\NCE}$ ($ \hat{c}-c $) }  & Learning rate & 10$^ {-3}$ & 10$^ {-3}$ & 10$^ {-3}$ \\[0.5em]
		&  Epochs & 3 & 3 & 3\\
		\midrule 
		\multirow{ 2}{*}{$\mathcal{L}_{\NCE}$ ($2\hat{c}-c $) }  & Learning rate & 10$^ {-3}$ & 10$^ {-3}$ & 10$^ {-3}$\\[0.5em]
		&  Epochs & 2 & 2 & 4\\
% 		\midrule 
% 		\multirow{ 2}{*}{$\mathcal{L}_{ReLU}$ ($ \hat{c}$) }  & Learning rate & 10$^ {-3}$ & 10$^ {-3}$ & 10$^ {-3}$ \\[0.5em]
% 		&  Epochs & 15 & 15 & 15\\
% 		\midrule 
% 		\multirow{ 2}{*}{$\mathcal{L}_{ReLU}$ ($ \hat{c}-c $) }  & Learning rate & 10$^ {-3}$ & 10$^ {-3}$ & 10$^ {-3}$ \\[0.5em]
% 		&  Epochs & 15 & 15 & 15\\
% 		\midrule 
% 		\multirow{ 2}{*}{$\mathcal{L}_{ReLU}$ ($2\hat{c}-c $) }  & Learning rate & $5\times10^ {-3}$ & $5\times10^ {-3}$ & $5\times10^ {-3}$\\[0.5em]
% 		&  Epochs & 15 & 15 & 15\\
		\midrule 
		\multirow{ 2}{*}{$\mathcal{L}_{\MAP}$ ($ \hat{c}$) }  & Learning rate & $5\times10^ {-3}$ & $5\times10^ {-3}$ & $5\times10^ {-3}$1\\[0.5em]
		&  Epochs & 5 & 5 & 5\\
		\midrule 
		\multirow{ 2}{*}{$\mathcal{L}_{\MAP}$ ($ \hat{c}-c $) }  & Learning rate & $5\times10^ {-3}$ & $5\times10^ {-3}$ & $5\times10^ {-3}$\\[0.5em]
		&  Epochs & 2 & 5 & 4\\
		\midrule 
		\multirow{ 2}{*}{$\mathcal{L}_{MAP}$ ($2\hat{c}-c $) }  & Learning rate & $5\times10^ {-3}$ & $5\times10^ {-3}$ & $5\times10^ {-3}$\\[0.5em]
		&  Epochs & 4 & 2 & 4\\
		\bottomrule
	\end{tabular}
	\caption{Choice of Hyperparameters (Diverse Bipartite Matching Problem) }
	% \label{table:hyperparams}
\end{table*}
%}
%\end{supertabular}
% The file named.bst is a bibliography style file for BibTeX 0.99c
\end{document}